\pdfoutput=1

\documentclass[11pt]{article}

\usepackage{acl}

\usepackage{times}
\usepackage{latexsym}

\usepackage[T1]{fontenc}

\usepackage[utf8]{inputenc}

\usepackage{microtype}

\usepackage{inconsolata}

\usepackage{graphicx}
\usepackage{amsmath}
\usepackage{amsthm}
\usepackage{booktabs}
\usepackage{multicol}
\usepackage{adjustbox}
\usepackage{subcaption}
\usepackage{multirow}
\usepackage{algpseudocode}
\usepackage{algorithm}
\usepackage{amssymb}
\usepackage{makecell}
\usepackage[multiple]{footmisc}
\usepackage{resizegather}
\usepackage{microtype}
\usepackage{bbm}
\usepackage{enumitem}
\usepackage{hyperref}
\usepackage{dsfont}

\DeclareMathOperator*{\argmax}{arg\,max}

\title{Re-evaluating Automatic LLM System Ranking for Alignment \\ with Human Preference} 

\author{Mingqi Gao$^{*1,2}$\qquad Yixin Liu$^{*2}$\qquad Xinyu Hu$^{1}$\qquad\\ \textbf{Xiaojun Wan}$^{1}$\qquad \textbf{Jonathan Bragg}$^{3}$\qquad \textbf{Arman Cohan}$^{2,3}$\vspace{6pt}\\
         $^{1}$Peking University\quad $^{2}$Yale University\quad $^{3}$Allen Institute for AI\vspace{4pt}\\
         \texttt{\{mingqi.gao,yixin.liu,arman.cohan\}@yale.edu}}

\begin{document}
\maketitle
\def\thefootnote{*}\footnotetext{Equal contribution.}\def\thefootnote{\arabic{footnote}}

\begin{abstract}
Evaluating and ranking the capabilities of different LLMs is crucial for understanding their performance and alignment with human preferences. Due to the high cost and time-consuming nature of human evaluations, an automatic LLM \textit{\textbf{bencher}} (i.e., an automatic evaluation framework that aims to rank LLMs based on their alignment with human preferences) is indispensable. An automatic LLM bencher consists of four components: the input set (e.g., a user instruction), the evaluation model (e.g., an LLM), the evaluation type (e.g., pairwise comparison), and the aggregation method (e.g., the ELO rating system). However, previous work has not thoroughly explored how to select these components or how their different combinations influence the results.
In this work, through controlled experiments, we provide a series of recommendations on how to choose each component to better automate the evaluation of LLMs. 
Furthermore, we discovered that when evaluating LLMs with similar performance, the performance of the automatic LLM bencher declines sharply, underscoring the limitations of current benchers and calling for future work.
Lastly, we found that the evaluation models' performance at the instance level (e.g., the accuracy of selecting the best output) does not always align with their effectiveness when used as a component of a bencher, highlighting the importance of dedicated system-level evaluation of benchers.
\end{abstract}

\section{Introduction}

Recently, various large language models (LLMs) \citep{DBLP:conf/nips/Ouyang0JAWMZASR22} and LLM-based agents \citep{DBLP:journals/fcsc/WangMFZYZCTCLZWW24} have been released, demonstrating strong capabilities across various tasks. These systems enable efficient interactions with human users and can be instructed to perform various complex activities, making their performance in such tasks an important aspect of evaluation. To benchmark the capabilities of these systems, human judgments of output quality remain indispensable as the gold standard, as many tasks do not have standard answers and are inherently open-ended \citep{DBLP:conf/nips/ZhengC00WZL0LXZ23,DBLP:conf/nips/DuboisLTZGBGLH23}.

Chatbot Arena \citep{DBLP:conf/icml/ChiangZ0ALLZ0JG24} embodies this concept. It is a real-time evaluation platform aimed at a large user base, where users can freely provide input, select any two hosted LLMs to generate responses, and indicate which one they prefer. Chatbot Arena periodically derives a leaderboard of various LLMs by aggregating instance-level pairwise human evaluations. 
To date, Chatbot Arena has collected over 1.5 million human judgments, involving more than 100 systems. 
Due to its substantial size and the comprehensiveness of the systems included, its LLM ranking has been widely regarded as a trustworthy indicator of an LLM’s general capabilities \citep{alpaca_eval, DBLP:journals/corr/abs-2406-11939, DBLP:journals/corr/abs-2406-04770,DBLP:journals/corr/abs-2405-20267}.

However, automatic LLM system rankers (i.e. benchers) are often needed because of the expensive and time-consuming nature of human evaluation.\footnote{
LLMs used to evaluate the output of other LLMs are also referred as ``evaluators'' or ``LLM-as-a-judge'' in the literature. We use the word ``bencher'' to explicitly emphasize the broader automatic evaluation frameworks used to derive rankings of models in a benchmark (in which an LLM evaluator is often used as part of the framework).}
As a result, several widely used automatic benchers have been proposed, such as Alpaca Eval \citep{alpaca_eval,DBLP:journals/corr/abs-2404-04475} and Arena Hard \citep{DBLP:journals/corr/abs-2406-11939}, which use a strong LLM such as GPT-4 \citep{DBLP:journals/corr/abs-2303-08774} to compare various systems on a carefully designed input set. 
These automatic benchers can produce LLM rankings that correlate well with Chatbot Arena, achieving Spearman's $\rho$ of around 0.95.
Therefore, they have been widely used to develop and evaluate LLMs for tasks like alignment fine-tuning \citep{tunstall2024zephyr, yang2024qwen2}.

Formally, an automatic LLM bencher for evaluating instruction following capabilities consists of the following components:
\begin{itemize}[topsep=0pt, align=left, leftmargin=0pt, labelindent=6pt,
listparindent=\parindent, labelwidth=0pt, itemindent=!, itemsep=0pt, parsep=0pt]
    \item \textit{Input Set}: A set of instructions that each LLM is tasked with to generate responses.
    \item \textit{Evaluation Model}: An (often strong) LLM  (e.g. GPT-4o) that serves as a proxy for humans in evaluating a model's responses.
    \item \textit{Evaluation Type}: The type of instance-level evaluation generated by the evaluation model, such as pointwise scores or pairwise comparisons, usually determined by the prompt.\footnote{Following the literature, throughout this paper, ``Instance-level'' refers to an evaluation instance, which could consist of a pair of model outputs (in pairwise evaluation) or a single model output (in pointwise evaluation).}
    \item \textit{Aggregation Method}: The method by which instance-level evaluations are converted into system (or model) scores and rankings. 
\end{itemize}

While these automatic benchers have shown a high level of alignment with human evaluations in Chatbot Arena, we note a lack of a systematic evaluation across all the above-mentioned components involved with the benchers. For example, while Arena Hard claims a higher alignment with human evaluation than Alpaca Eval and attributes its success to the careful selection of its input set, they also adopt different evaluation types for the evaluation: Arena Hard performs five-point pairwise comparisons, while Alpaca Eval produces binary comparisons.
Therefore, the exact source of its superior performance remains unclear because of the lack of controlled comparisons.

Therefore, we aim to conduct a more rigorous examination of these automatic benchers.
To this end, the first research question we explore is \textbf{RQ1: how to choose the appropriate components for building an effective automatic LLM bencher?}
Specifically, we perform controlled comparisons of various input sets, evaluation models, evaluation types, and aggregation methods to investigate which choices of each component maximize the bencher's performance.
Our key findings are:

\textit{(1) Input Set}: Compared to the earlier released Alpaca Eval, using Arena Hard as the input set, which is filtered from crowdsourced user prompts through an LLM-powered pipeline, always yields higher correlations with the system rankings of Chatbot Arena.
    
\textit{(2) Evaluation Model}: While proprietary models like GPT-4-turbo generally perform best as evaluation models, the open-source Llama-3.1-70B~\cite{dubey2024llama} also delivers strong performance, particularly when paired with a suitable combination of other components in automatic benchers.
    
\textit{(3) Evaluation Type}: With a fixed number of LLM queries, pointwise evaluation is slightly better than base pairwise evaluation\footnote{We refer to binary pairwise (win/lose) comparisons as the base pairwise evaluation method.} for strong LLMs, but pairwise evaluation considerably outperforms pointwise with less capable LLMs. Notably, in almost all cases, the widely used 5-point pairwise evaluation performs \textit{worse} than the base pairwise approach. Additionally, using reference systems is an efficient strategy.

\textit{(4) Aggregation Method}:
The system rankings produced by the Bradley-Terry model \citep{10.1093/biomet/39.3-4.324} and win ratio are mostly the same when the evaluation type is base pairwise. Interestingly, when the evaluation type is pointwise, the arithmetic mean, the most common aggregation method, is not always optimal.

Having provided a thorough evaluation of the automatic benchers, we take a step forward to explore a more challenging setting.
A critical question regarding the evaluation of the LLM benchers is its reliability when the task is to evaluate similar-performing LLMs, which is an important use case. For example, these benchers are widely used to compare different model checkpoints fine-tuned from the same pre-trained model, which have similar performance and behavior. Finally, based on the above experimental results, it seems that we can easily build an effective automatic bencher using GPT-4-turbo as the evaluation model, achieving a Spearman's rho of over 0.90 with Chatbot Arena. However, \textbf{does the performance of the bencher decline when systems with smaller performance differences are selected for evaluation (RQ2)?}

One aim of our above analyses was to identify suitable evaluation models (see Finding (2) above). Related work like LLMBAR \citep{DBLP:conf/iclr/ZengYG0G024} and RewardBench \citep{DBLP:journals/corr/abs-2403-13787} has taken a different approach to evaluating evaluation models, by measuring correlation or accuracy with ground truth instance-level human annotations rather than aggregate system rankings from Chatbot Arena. Compared to the crowd-sourced ChatBot Arena, human annotations collected for instance-level evaluations are usually of higher quality but lack the scale in terms of the number of systems used to generate evaluation examples. Sometimes system information is even unavailable for instance-level evaluation datasets.
Given the significant efforts made in this direction, it would be desirable for instance-level evaluation results to also generalize to system-level evaluation, which leads to our \textbf{RQ3: can we use instance-level rankings of evaluation models as a good reference to select evaluation models for LLM benchers?}

Our investigation of the three research questions underscores several key findings for the users and developers of LLM benchers:
(1) RQ1: It is important to choose the suitable components to build a bencher. For example, we find that using Llama-3.1-70B with a pairwise comparison protocol, the BT model for aggregation on Arena Hard can yield a comparable Spearman correlation of 0.93 with human annotators compared to the standard AlpacaEval2 evaluation pipeline where GPT-4 is used as the evaluator.
(2) RQ2: Despite their promising performance in general, the benchers' alignment with humans degrades quickly when they are used to rank close-performing systems, calling for further work to enhance their robustness and accuracy.
(3) RQ3: The evaluation results of LLM-based evaluators on the instance level do not always generalize to the system-level performance of the corresponding benchers. This highlights the need for dedicated evaluations at the system level.

\section{Problem Formalization}

In this section, we formalize some crucial concepts and our research questions.

\subsection{Automatic LLM Bencher}
Given a set of systems ($\mathcal{S} = \{s_1, s_2, \dots, s_n\}$) to be evaluated/ranked , an automatic bencher $\mathcal{J}_A$ is composed of the following four components:

\noindent \textbf{Input Set} $\mathcal{X} = \{x_1, x_2, \dots, x_m\}$. Each system $s_i$ generates a response $o_{ij}$ for each input $x_j$.

\noindent \textbf{Evaluation Model} $\mathcal{E}$: This is an LLM (e.g., GPT-4-turbo) that acts as a proxy for human evaluators. It generates instance-level evaluations for each system.  The evaluation model $\mathcal{E}$ outputs instance-level evaluation for each response $o_{ij}$ .

\noindent \textbf{Evaluation Type} $\mathcal{T}$: This refers to the type of evaluation generated at the instance level. There are two main types\footnote{There is also the "list-wise" approach, where a list of model outputs is provided, and the LLM is tasked with ranking them at once. However, this type of evaluation is not commonly employed in LLM-based judges or benchmarks.}:

\noindent (1) Pointwise Evaluation: The evaluation model assigns a score to a system’s response. The pointwise score of the $i$-th system's response to the $j$-th input is denoted as $\mathcal{T}_{\text{pointwise}}(\mathcal{E}, x_j, o_{ij}) \in \mathbb{R}$. The total number of instance-level evaluations is $mn$.

\noindent (2) Pairwise Evaluation: the evaluation model compares the quality of two responses $o_{ij}, o_{kj}$ generated by two systems $s_i$ and $s_k$ on the same input $x_j$. The base comparison result is binary, indicating whether $s_i$ performed better than $s_k$ or vice versa. This can be expressed as: $\mathcal{T}_{\text{base.pairwise}}(\mathcal{E}, x_j, o_{ij}, o_{kj}) \in \{o_{ij} > o_{kj}, o_{ij} < o_{kj}\}$. In 5-point pairwise comparisons, the evaluation model provides more detailed feedback about the difference between two systems' performances, capturing both the magnitude and direction of the preference. The 5 possible comparison outcomes for systems $s_i$ and $s_k$ on input $x_j$ are: $
\mathcal{T}_{\text{5-point.pairwise}}(\mathcal{E}, x_j, o_{ij}, o_{kj}) \in \{o_{ij} \ll o_{kj}, o_{ij} < o_{kj}, o_{ij} = o_{kj}, o_{ij} > o_{kj}, o_{ij} \gg o_{kj}\}
$. The total number of instance-level pairwise evaluations is $mn(n-1)$ rather than $mn(n-1)/2$ because we always need to swap the presentation order of the responses $s_i$ and $s_j$ to address the position bias of LLM-based evaluations. In addition, due to the high computational cost of performing full-scale comparisons, some automatic LLM benchers \citep{alpaca_eval,DBLP:journals/corr/abs-2406-11939} introduce a \textbf{reference system} in pairwise evaluation. In this approach, all systems under evaluation only need to be compared with a reference system (e.g. GPT-4), which reduces the complexity of pairwise evaluation to a level similar to that of pointwise evaluation.

\noindent \textbf{Aggregation Method} $\mathcal{G}$: This will create a mapping from systems to scores ($\mathcal{G}(s_i)\in \mathbb{R}$), depending on the evaluation type.

For pointwise evaluations, aggregation can be done via the mean or median of the instance-level scores across all inputs:
\begin{align*}
\mathcal{G}_{\text{mean}}(s_i) = & \frac{1}{m} \sum_{j=1}^{m} \mathcal{T}_{\text{pointwise}}(\mathcal{E}, x_j, o_{ij}) \\
\mathcal{G}_{\text{median}}(s_i) = & \text{median}\left(\{\mathcal{T}_{\text{pointwise}}(\mathcal{E}, x_j, o_{ij})\}_{j=1}^m \right)
\end{align*}

For pairwise evaluations, we consider the Bradley-Terry model \citep{10.1093/biomet/39.3-4.324} and win ratio\footnote{We do not consider the online ELO rating system \citep{elo1978rating} because the rankings it produces are influenced by the input order of instance-level evaluations.}. They can only be applied to base pairwise instance-level evaluations, so other types of instance-level evaluations need to be converted to the base pairwise type. The conversion rules are described in Appendix \ref{appendix:conversion}. The Bradley-Terry model $\mathcal{G}_{\text{BT}}$ computes the strength of each system via maximum likelihood estimate, and the win ratio is the proportion of pairwise wins for a system over others: 

{\small
\begin{align*}
& \mathcal{G}_{\text{win\_ratio}}(s_i) = \\
& \frac{\sum_{j=1}^{m} \sum_{k \neq i} \mathds{1}(\mathcal{T}_{\text{base.pairwise}}(\mathcal{E}, x_j, o_{ij}, o_{kj}) = o_{ij} > o_{kj})}{\sum_{j=1}^{m} \sum_{k \neq i} 1} 
\end{align*}
}

where $m$ is the total number of inputs and the denominator counts the total number of pairwise comparisons involving $s_i$. Since pointwise data can be converted into pairwise data, all aggregation methods for pairwise evaluations can also be applied to pointwise evaluations after conversion.

\subsection{How to Evaluate a Bencher}

\noindent \textbf{Ground Truth.} From Chatbot Arena, we obtain system ratings, denoted as $r(s_i)$, for each system $s_i \in S$, where $r(s_i) \in \mathbb{R}$ represents the overall performance of the system as derived from human judgments. The ranking of the systems is obtained by sorting them based on their ratings: $R_H: S \to \{1, 2, \dots, n\}$, where $R_H(s_i)$ represents the rank of system $s_i$ based on the ratings $r(s_i)$.

\noindent \textbf{Evaluation Measure.} The performance of an automatic bencher $\mathcal{J}_A$ is evaluated by comparing its ranking $R_A$ of systems with the ground truth ranking $R_H$, using Spearman’s rank correlation coefficient ($\rho(R_A, R_H)$) and Kendall's tau ($\tau(R_A, R_H)$). $\rho$ is more widely used.

\noindent \textbf{Controllable Kendall’s Tau} ($\tau_u$): Inspired by \citet{DBLP:conf/naacl/DeutschDR22}, we propose controllable Kendall’s tau that evaluates the agreement between rankings by focusing only on system pairs where the performance difference is small, i.e., where the absolute difference in the human-provided scores for two systems is less than a specified threshold $u$. This is useful for evaluating the bencher's ability to distinguish between closely matched systems.

Let $\Delta r(s_i, s_k) = |r(s_i) - r(s_k)|$ represent the performance difference between systems $s_i$ and $s_k$, where $r(s_i)$ is the system ratings derived from Chatbot Arena. For a given threshold $u$, we define the set of system pairs that are considered for evaluation as: $
P_{u} = \{(s_i, s_k) \mid s_i, s_k \in \mathcal{S}, \Delta r(s_i, s_k) \leq u, \, i \neq k \}$. To prevent the performance gap between two systems from being so small that humans also find it difficult to distinguish between them, we only consider system pairs whose 95\% confidence intervals in the Chatbot Arena ratings do not overlap: $
Q = \{(s_i, s_k) \mid s_i, s_k \in \mathcal{S}, \text{CI}(r(s_i)) \cap \text{CI}(r(s_k)) = \varnothing, i \neq k \}$.
The controllable Kendall’s tau $\tau_u$ is then calculated only using the system pairs in $P_{u}\cap Q$:

{\small
\begin{align*}
& \tau_u(R_A, R_H) = \frac{C_u - D_u}{\sqrt{(C_u + D_u + T_{A,u})(C_u + D_u + T_{H,u})}}
\end{align*}
}

where $C_u$ is the number of concordant pairs (i.e., pairs of systems $(s_i, s_k)$ where the rank order between $s_i$ and $s_k$ is the same in both $R_A$ and $R_H$). $D_u$ is the number of discordant pairs. $T_{A,u}$ and $T_{H,u}$ are the number of ties in the automatic bencher’s ranking $R_A$ and the human-provided ranking $R_H$. Only the system pairs within the set $P_{u}$ are used.

\subsection{Different Ways of Selecting Evaluation Models}

We are comparing different evaluation models $\mathcal{E}$ based on their ability to rank systems or predict instance-level human preferences. There are three key settings for evaluating evaluation models:

\paragraph{Setting 1: Standard Meta-Evaluation with System-Level Rankings ($R_{\mathcal{E}}^{(1)}$)}. The idea of Setting 1 is to rank evaluation models based on how well their corresponding automatic benchers align with the system rankings derived from Chatbot Arena. It is the standard way benchmarks such as Arena Hard are designed.

An automatic bencher $\mathcal{J}_A^{(i)}$ is composed of an evaluation model $\mathcal{E}_i$ and \textbf{fixed} input set $\mathcal{X}$, evaluation type $\mathcal{T}$, and aggregation method $\mathcal{G}$, producing a system ranking $R_A^{(i)}$. The performance of evaluation model $\mathcal{E}_i$ is measured by comparing $R_A^{(i)}$ with $R_H$ using either Spearman’s $\rho(R_A^{(i)}, R_H)$, or Kendall’s $\tau(R_A^{(i)}, R_H)$ mentioned above. Evaluation models are ranked based on their performance, leading to a ranking for evaluation models: $
R_{\mathcal{E}}^{(1)}: \{\mathcal{E}_1, \mathcal{E}_2, \dots, \mathcal{E}_K\} \to \{1, 2, \dots, K\}$.

\paragraph{Setting 2: Instance-Level Human Judgments as Ground Truth ($R_{\mathcal{E}}^{(2)}$).} We consider instance-level human judgments as the ground truth, without system-level aggregation or knowledge of which systems generated the responses.

Let $\mathcal{D} = \{(x_j, o_j^{(1)}, o_j^{(2)}, h_j)\}_{j=1}^m$ represent the dataset, where each entry consists of an input $x_j$, two responses $o_j^{(1)}$,$o_j^{(2)}$, and human preference $h_j \in \{o_j^{(1)} > o_j^{(2)}, o_j^{(1)} < o_j^{(2)}\}$, indicating which response was preferred.

The evaluation model $\mathcal{E}_i$ generates predictions $\hat{h}_j^{(i)}$ = $\mathcal{T}_{\text{base.pairwise}}(\mathcal{E}_i, x_j, o_j^{(1)}, o_j^{(2)})$, and the instance-level accuracy is defined as: 
\begin{align*}
\text{Accuracy}(\mathcal{E}_i) = \frac{1}{m} \sum_{j=1}^{m} \mathds{1}(\hat{h}_j^{(i)} = h_j)
\end{align*}

Evaluation models are ranked ($R_{\mathcal{E}}^{(2)}$) based on their accuracy in predicting human preferences.

\paragraph{Setting 3: Instance-Level Human Judgments with System Information and Aggregation ($R_{\mathcal{E}}^{(3)}$).} We design this new setting to help us compare \textbf{Setting 1} and \textbf{Setting 2}. If we have access to instance-level human judgments as well as system information, we can aggregate both human judgments and evaluation model predictions into system-level rankings.

Dataset with system information: Each entry in the dataset $\mathcal{D}' = \{(x_j, o_j^{(1)}, o_j^{(2)}, s_j^{(1)}, s_j^{(2)}, h_j)\}_{j=1}^m$ includes: an input $x_j$, two responses $o_j^{(1)}$,$o_j^{(2)}$ generated by systems $s_j^{(1)}, s_j^{(2)}$, and human preference $h_j \in \{o_j^{(1)} > o_j^{(2)}, o_j^{(1)} < o_j^{(2)}\}$, indicating which response was preferred\footnote{Please note that the instance-level human judgments in Chatbot Arena cannot be utilized in Setting 3, as they are not fully disclosed. Specifically, $x_j$, $o_j^{(1)}$, and $o_j^{(2)}$ are unavailable, and only $s_j^{(1)}$, $s_j^{(2)}$, and $h_j$ are provided. Consequently, only the aggregated system rankings from Chatbot Arena (as used in Setting 1) are typically used.}. The steps to producing evaluation model rankings are as follows: (1) Human preferences are aggregated to produce a system ranking $R_H'$ with an aggregate method $\mathcal{G}$. (2) The evaluation model $\mathcal{E}_i$ provides instance-level predictions with $\mathcal{T}$, which are aggregated using the same method $\mathcal{G}$ to produce a system ranking $R_A^{(i)}$. (3) Compare the two system rankings $R_H'$ and $R_A^{(i)}$ using Spearman’s $\rho(R_A^{(i)}, R_H')$ or Kendall’s $\tau(R_A^{(i)}, R_H')$. (4) The evaluation models are then ranked ($R_{\mathcal{E}}^{(3)}$) based on the correlation between their aggregated rankings and the aggregated human judgments.

\textbf{Setting 1 and Setting 3 are system-level evaluation}, because aggregation methods are applied and a correlation value is calculated between two sets of system rankings, while \textbf{Setting 2 is the instance-level evaluation.} Note that we set the evaluation type to $\mathcal{T}_{\text{base.pairwise}}$ and aggregation method to $\mathcal{G}_{\text{BT}}$ in Setting 1 and Setting 3 to compare the results of the three settings more fairly.

\subsection{Research Questions}
Our research questions can be formalized as:

\begin{itemize}[topsep=0pt, align=left, leftmargin=0pt, labelindent=6pt,
listparindent=\parindent, labelwidth=0pt, itemindent=!, itemsep=0pt]
    \item RQ1: How to choose the appropriate components for building an effective automatic LLM bencher?
    
    $\Rightarrow\mathcal{X}, \mathcal{E}, \mathcal{T}, \mathcal{A} = \argmax_{\mathcal{X}, \mathcal{E}, \mathcal{T}, \mathcal{A}} \, \rho(R_A, R_H)$

    \item RQ2: Does the performance of the bencher decline when systems with smaller performance differences are selected for evaluation? 
    
    $\Rightarrow\text{Does } \tau_u(R_A, R_H) \text{ decrease}  \text{ as } u \text{ decrease} ?$
    
    \item RQ3: Can we use instance-level rankings of evaluation models as a good reference to select evaluation models for LLM benchers?
    
    $\Rightarrow \, \rho(R_{\mathcal{E}}^{(1)}, R_{\mathcal{E}}^{(2)}) \approx 1 \quad \text{and} \quad \rho(R_{\mathcal{E}}^{(1)}, R_{\mathcal{E}}^{(3)}) \approx 1 \quad \text{and} \quad \rho(R_{\mathcal{E}}^{(2)}, R_{\mathcal{E}}^{(3)}) \approx 1?$
\end{itemize}

\begin{table*}[t]
\small
\centering
\begin{tabular}{@{}llcccccccc@{}}
\toprule
\multicolumn{10}{c}{Alpaca Eval as the input set} \\ \midrule
\multicolumn{2}{c}{Evaluation Model} & \multicolumn{2}{c}{llama-3.1-70b} & \multicolumn{2}{c}{mixtral-8x7b} & \multicolumn{2}{c}{gpt-4o} & \multicolumn{2}{c}{gpt-4-turbo} \\ \midrule
Type & \multicolumn{1}{l}{Aggregation} & $\rho$ & $\tau$ & $\rho$ & $\tau$ & $\rho$ & $\tau$ & $\rho$ & $\tau$ \\ 
\midrule
pointwise & bradley\_terry & 0.8844 & 0.7255 & 0.4757 & 0.3333 & 0.8782 & 0.7255 & 0.8611 & 0.6951 \\
pointwise & win\_ratio & 0.8844 & 0.7255 & 0.4757 & 0.3333 & 0.8782 & 0.7255 & 0.8611 & 0.6951 \\
pointwise & mean & 0.5604 & 0.3987 & 0.2054 & 0.1373 & 0.8535 & 0.6993 & 0.8493 & 0.6863 \\
pointwise & median & 0.1806 & 0.1373 & 0.3953 & 0.2810 & 0.8762 & 0.7386 & 0.8803 & 0.7255 \\ \midrule
pairwise\_base & bradley\_terry & \textbf{0.9009} & \textbf{0.7647} & \textbf{0.5253} & \textbf{0.3856} & \textbf{0.8838} & \textbf{0.7475} & \textbf{0.9112} & 0.7778 \\
pairwise\_base & win\_ratio & \textbf{0.9009} & \textbf{0.7647} & \textbf{0.5253} & \textbf{0.3856} & \textbf{0.8838} & \textbf{0.7475} & \textbf{0.9112} & 0.7778 \\ \midrule
pairwise\_5point & bradley\_terry & 0.5026 & 0.3595 & 0.4510 & 0.3203 & 0.6615 & 0.5033 & 0.8184 & 0.6471 \\
pairwise\_5point & win\_ratio & 0.5046 & 0.3725 & 0.3251 & 0.2549 & 0.6037 & 0.4379 & 0.7977 & 0.6471 \\ \midrule
pairwise\_base\_ref & bradley\_terry & 0.8596 & 0.7124 & 0.4530 & 0.3333 & 0.8803 & 0.7124 & 0.9071 & \textbf{0.7908} \\
pairwise\_base\_ref & win\_ratio & 0.8617 & 0.6993 & 0.4530 & 0.3333 & 0.8741 & 0.6993 & 0.9092 & 0.7778 \\ \midrule
pairwise\_5point\_ref & bradley\_terry & 0.5005 & 0.3856 & 0.4551 & 0.3333 & 0.6636 & 0.5163 & 0.8658 & 0.7124 \\
pairwise\_5point\_ref & win\_ratio & 0.4881 & 0.3725 & 0.3457 & 0.2680 & 0.5707 & 0.4510 & 0.7090 & 0.5556 \\
\midrule
\multicolumn{10}{c}{Arena Hard as the input set} \\ \midrule
\multicolumn{2}{c}{Evaluation Model} & \multicolumn{2}{c}{llama-3.1-70b} & \multicolumn{2}{c}{mixtral-8x7b} & \multicolumn{2}{c}{gpt-4o} & \multicolumn{2}{c}{gpt-4-turbo} \\ \midrule
Type & \multicolumn{1}{l}{Aggregation} & $\rho$ & $\tau$ & $\rho$ & $\tau$ & $\rho$ & $\tau$ & $\rho$ & $\tau$ \\ 
\midrule
pointwise & bradley\_terry & 0.7207 & 0.5639 & 0.2693 & 0.2026 & 0.9401 & 0.8039 & 0.9092 & 0.7778 \\
pointwise & win\_ratio & 0.7207 & 0.5639 & 0.2693 & 0.2026 & 0.9401 & 0.8039 & 0.9092 & 0.7778 \\
pointwise & mean & 0.6863 & 0.4771 & 0.6739 & 0.5163 & \textbf{0.9628} & \textbf{0.8693} & \textbf{0.9463} & \textbf{0.8562} \\
pointwise & median & 0.5439 & 0.4248 & 0.1538 & 0.1111 & 0.9587 & 0.8562 & 0.9319 & 0.7908 \\ \midrule
pairwise\_base & bradley\_terry & \textbf{0.9340} & \textbf{0.8170} & \textbf{0.7441} & 0.5948 & 0.9443 & 0.8039 & 0.9381 & 0.8431 \\
pairwise\_base & win\_ratio & \textbf{0.9340} & \textbf{0.8170} & \textbf{0.7441} & 0.5948 & 0.9443 & 0.8039 & 0.9381 & 0.8431 \\ \midrule
pairwise\_5point & bradley\_terry & 0.8411 & 0.7124 & 0.7296 & \textbf{0.6078} & 0.9216 & 0.7778 & \textbf{0.9463} & 0.8431 \\
pairwise\_5point & win\_ratio & 0.8142 & 0.6732 & 0.6223 & 0.4902 & 0.9360 & 0.8039 & 0.9133 & 0.8039 \\ \midrule
pairwise\_base\_ref & bradley\_terry & 0.8989 & 0.7647 & 0.7372 & 0.6033 & 0.9530 & 0.8262 & 0.9278 & 0.8170 \\
pairwise\_base\_ref & win\_ratio & 0.8968 & 0.7516 & 0.7372 & 0.6033 & 0.9427 & 0.8000 & 0.9257 & 0.8039 \\ \midrule
pairwise\_5point\_ref & bradley\_terry & 0.8101 & 0.6732 & 0.7523 & 0.6340 & 0.9154 & 0.7516 & 0.8885 & 0.7778 \\
pairwise\_5point\_ref & win\_ratio & 0.7131 & 0.5817 & 0.6367 & 0.4771 & 0.8473 & 0.6863 & 0.7915 & 0.6209 \\
\midrule
\end{tabular}
\vspace{-2mm}
\caption{Standard meta-evaluation results of different combinations of input sets, evaluation models, evaluation types, and aggregation methods (the correlation between the automatic benchers' and ChatBot Arena's evaluations). For the pairwise evaluation using a reference system, we show the results with gpt-4-0314 as the reference system. For the results with other LLMs as evaluation models, please refer to Table \ref{tab:type_aggregation_main-2} and Table \ref{tab:type_aggregation_main-3} in appendix.}
\label{tab:type_aggregation_main}
\vspace{-3mm}
\end{table*}

\section{Experiments and Analyses}

\subsection{Setups}
We selected two input sets, Arena Hard \citep{DBLP:journals/corr/abs-2406-11939} and Alpaca Eval \citep{alpaca_eval}, 18 LLMs as the systems to be evaluated, and 12 LLMs, including four proprietary OpenAI models and six open-source models, to serve as evaluation models. Please see Appendix \ref{appendix:experimental_setup} for detailed information.

\subsection{RQ1: How to choose \texorpdfstring{$\mathcal{X}, \mathcal{E}, \mathcal{T}, \mathcal{A}$}{X, E, T, A} to maximize the bencher’s performance?}
This section discusses component selection for the bencher and includes a cost analysis.

\paragraph{Input Set.} Table \ref{tab:type_aggregation_main} confirms that using Arena Hard as input set always yields higher correlations with the system rankings of Chatbot Arena compared to Alpaca Eval through different combinations of $\mathcal{E}, \mathcal{T}, \mathcal{A}$.

\paragraph{Evaluation Type.} Table \ref{tab:type_aggregation_main} shows the performance of a bencher under different combinations of evaluation types and aggregation methods, with several representative LLMs as evaluation models. For many open-source models with fewer parameters, base pairwise is the optimal evaluation type, offering a significant advantage over other evaluation types. For strong proprietary models, the difference between base pairwise and pointwise evaluation types is small. Across all models used as evaluation models, 5-point pairwise generally performs worse than base pairwise, with the decline being particularly pronounced for open-source models, possibly due to their weaker ability to follow complex instructions. \textbf{Therefore, using 5-point pairwise as the evaluation type, as in Arena Hard, may be problematic.} In addition, by comparing the results of pairwise evaluation using the full data with those using GPT-4 as the reference system, we found that for most strong evaluation models, the performance of the bencher is similar when using the reference system, indicating that this is an efficient strategy. 
Further discussion on the reference system is in Appendix \ref{appendix:reference_system}.

\begin{figure*}[t]
  \includegraphics[width=0.95\linewidth]{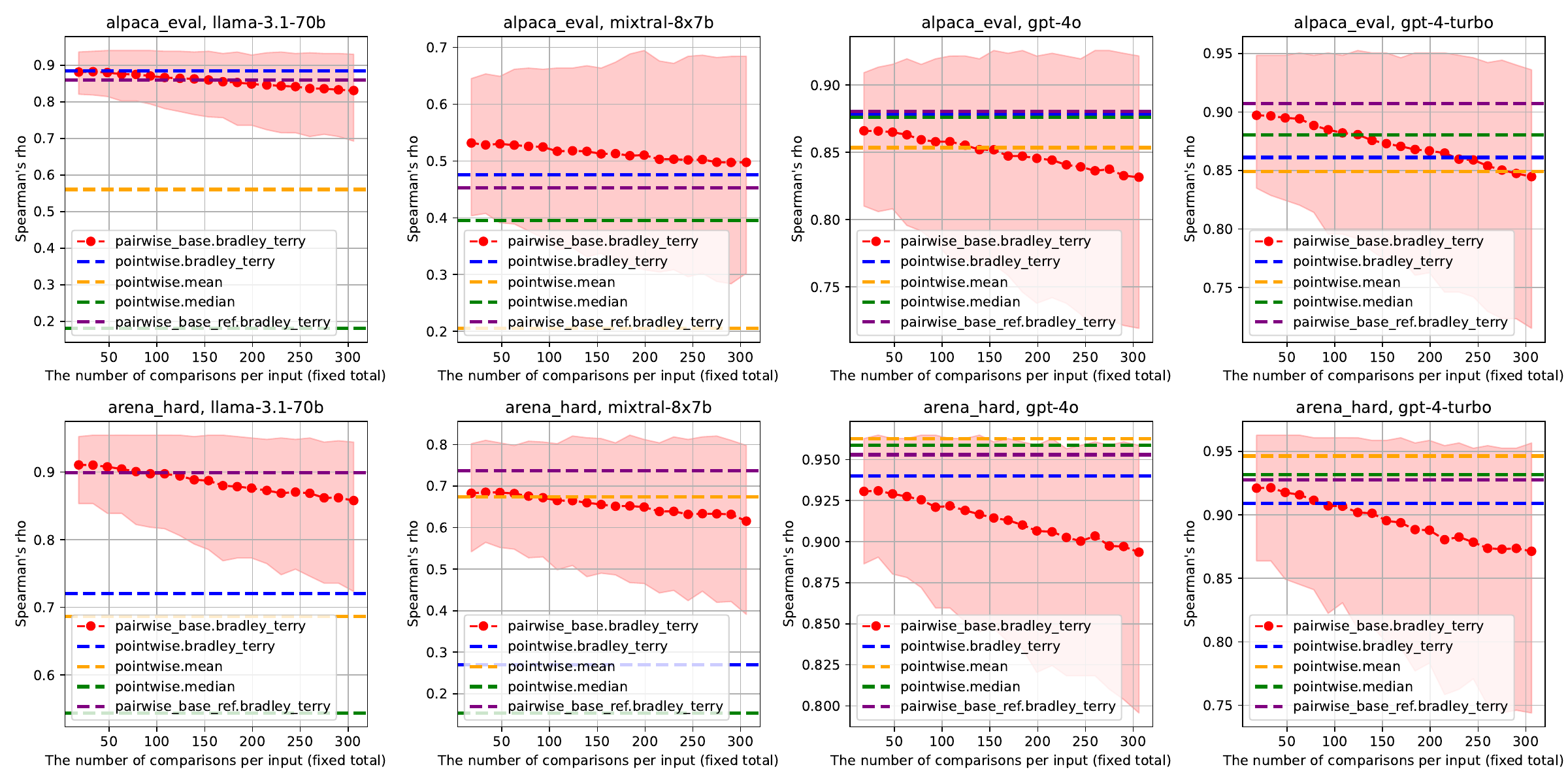}
    \vspace{-3mm}
  \caption{Bootstrapping mean and 95\% confidence interval of the correlations between the system rankings of automatic pairwise evaluation and human judgment when the total number of examples is set equal to that of pointwise evaluation. Results for pointwise and pairwise evaluations using gpt-4-0314 as the reference system are shown as horizontal lines. Both input sets and comparisons are sampled with replacement, with 1000 iterations. See Figure \ref{fig:p_pairs_sampling2} and \ref{fig:p_pairs_sampling3} in the appendix for other evaluation models.}
    \label{fig:p_pairs_sampling}
    \vspace{-2mm}
\end{figure*}

\paragraph{Aggregation Method.} As shown in Table \ref{tab:type_aggregation_main}, when the evaluation type is base pairwise, the system rankings produced by the Bradley-Terry model and win ratio aggregation methods are mostly the same. Interestingly, when the evaluation type is pointwise, mean, the most common aggregation method, is not always optimal. Moreover, when the evaluation model is a smaller open-source model, pointwise combined with the Bradley-Terry model significantly outperforms its combination with mean. After examination, we found that this is because the majority of instance-level scores produced by these evaluation models are very similar, and aggregating them using mean or median across all samples leads to system scores that are hardly effective. In contrast, converting these scores to base pairwise retains the slight differences in scores for different responses to the same input. On the other hand, for OpenAI models, the performance of aggregating pointwise type data using either mean or the Bradley-Terry model is comparable.

\begin{table}[t]
\resizebox{\linewidth}{!}{
\begin{tabular}{@{}lcccc@{}}
\toprule
 & \multicolumn{2}{l}{Alpaca Eval} & \multicolumn{2}{l}{Arena Hard} \\ \midrule
Evaluation Model & $\rho$ & Ranking & $\rho$ & Ranking \\ \midrule
gpt-4-turbo & 0.9112 & 1 & 0.9381 & 2 \\
gpt-4o & 0.8838 & 3 & 0.9443 & 1 \\
gpt-4o-mini & 0.6821 & 4 & 0.8101 & 5 \\
gpt-3.5-turbo & 0.4923 & 7 & 0.6636 & 8 \\
llama-3.1-70b & 0.9009 & 2 & 0.9340 & 3 \\
qwen-2.5-72b & 0.4861 & 8 & 0.8411 & 4 \\
mixtral-8x7b & 0.5253 & 6 & 0.7441 & 6 \\
llama-3.1-8b & 0.6347 & 5 & 0.6656 & 7 \\
glm-4-9b & 0.2755 & 9 & 0.6367 & 9 \\
mistral-7b-v0.3 & 0.2549 & 10 & 0.4076 & 10 \\
mistral-7b-v0.1 & 0.1476 & 11 & 0.2755 & 12 \\
llama-2-7b & -0.1291 & 12 & 0.2839 & 11 \\ \bottomrule
\end{tabular}
}
\vspace{-3mm}
\caption{Evaluation model rankings under Setting 1. The evaluation type and aggregation method are fixed to base pairwise and the Bradley-Terry model for all evaluation models.}
\label{tab:evaluation model_setting1}
\vspace{-5mm}
\end{table}

\noindent \textbf{Evaluation Models.} Table \ref{tab:evaluation model_setting1} displays the performance of benchers with different LLMs as evaluation models. We found that gpt-4-turbo and gpt-4o are the strongest evaluation models. The rankings of evaluation models obtained from the Alpaca Eval and Arena Hard input sets show slight differences, but the correlation between them is high, with $\rho=0.98$. Additionally, most of the smaller open-source models perform unsatisfactorily as evaluation models. However, llama-3.1-70b is an exception, performing even better than gpt-3.5-turbo and gpt-4o-mini.

\noindent \paragraph{Cost Analysis.} In the previous analysis, we did not take into account the total number of instance-level evaluations produced by the evaluation models. As mentioned earlier, the total number of instance-level evaluations of pointwise evaluation and pairwise evaluation are different: pointwise evaluations amount to $mn$, whereas pairwise evaluations amount to $mn(n-1)$. This implies that using pairwise as the evaluation type will incur significantly higher costs, especially when using proprietary models as evaluation models. Therefore, we consider it necessary to limit the number of instance-level evaluations for pairwise to be equal to that of pointwise, i.e., $mn$. In this scenario, when using pairwise as the evaluation type, we need to balance the \textbf{number of inputs} with the \textbf{number of comparisons per input}. Figure \ref{fig:p_pairs_sampling} illustrates the bootstrapping mean and confidence intervals for base pairwise as the evaluation type under different numbers of comparisons per input. We found that:

1. When some open-source models are used as evaluation models, base pairwise still has a certain advantage even when the total number of instance-level evaluations for base pairwise is limited to that of the pointwise type. 

\begin{figure*}[t]
  \includegraphics[width=\linewidth]{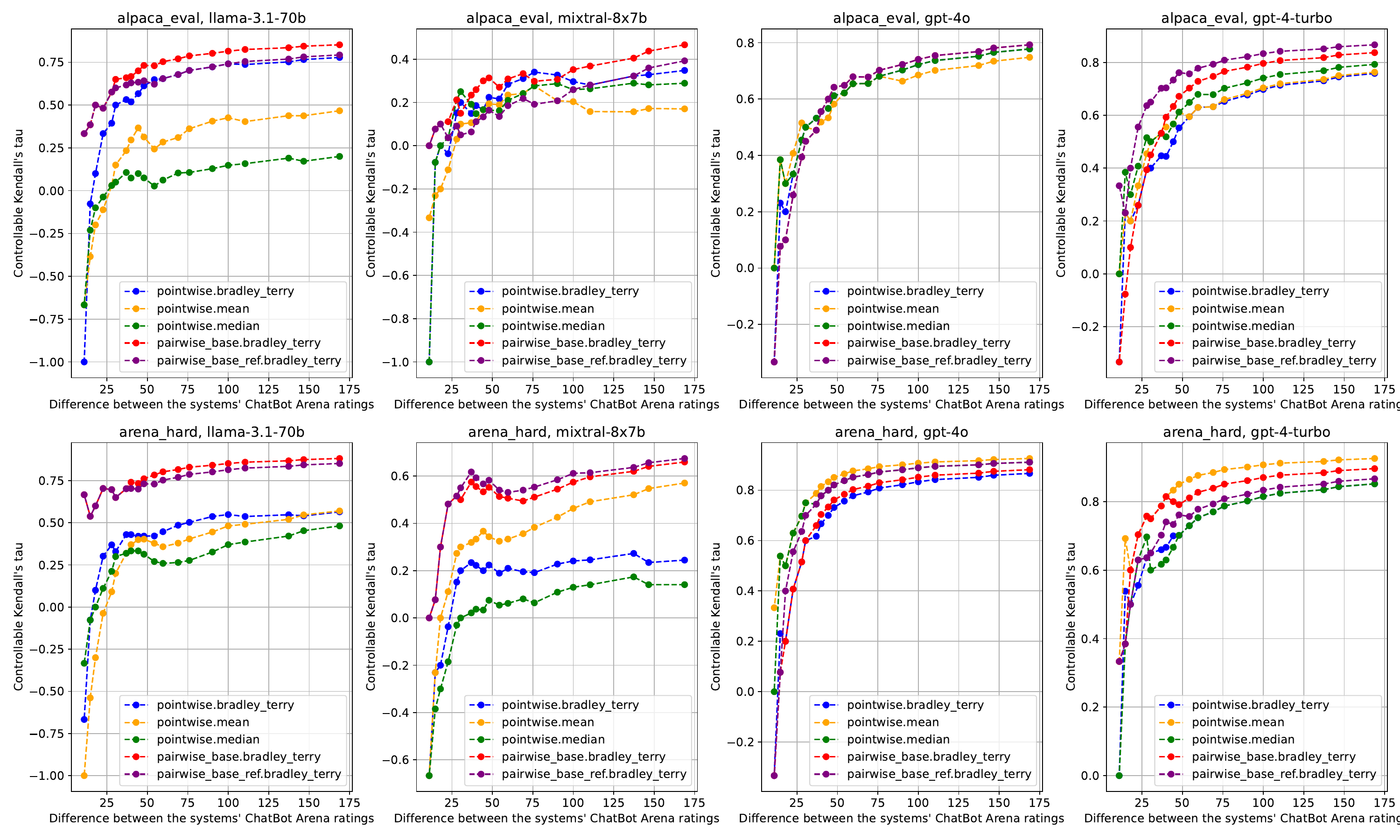}
  \caption{Controllable Kendall's tau ($\tau_u$) between the system rankings from automatic LLM benchers and human judgment when only partial system pairs are used. The X-axis denotes the value of threshold $u$, which controls the maximum difference between the systems' ChatBot Arena ratings. For \texttt{pairwise\_base\_ref}, we show the results with gpt-4-0314 as the reference system. Across all settings, we found that the benchers' performance degrades when they evaluate close-performing systems. See Figure \ref{fig:controllable_kendall_tau_2} and \ref{fig:controllable_kendall_tau_3} in the appendix for other evaluation models.}
    \label{fig:controllable_kendall_tau}
\end{figure*}

2. As the number of comparisons per input increases (i.e., the number of inputs decreases), both the bootstrapping mean and the lower bound of the confidence interval for bencher performance decline. \textbf{This suggests that we should prioritize using more inputs rather than conducting more comparisons per input.} As mentioned earlier, using a reference system is a practical strategy. For a more detailed cost analysis, please see appendix \ref{appendix:cost_analysis}.

\subsection{RQ2: Does the performance of automatic LLM benchers degrade when evaluating LLMs with similar performance?}

As shown in Figure \ref{fig:chatbot_arena_rating} in the appendix, the performance differences among the selected LLM systems are uneven, so the difficulty of distinguishing between different LLMs varies. We introduced a threshold $u$ so that only system pairs with performance differences smaller than it are used to calculate $\tau_u$. Since fewer system pairs meet the requirement as $u$ decreases, we can control the threshold $u$ to select a specific proportion of system pairs. Specifically, we selected 5\%, 10\%, ..., and 100\% of system pairs and observed the changes in bencher performance. Figure \ref{fig:controllable_kendall_tau} shows that as $u$ decreases (i.e., the performance differences between systems become smaller), the performance of almost all benchers declines sharply. For example, when Alpaca Eval is used as the input set and gpt-4o is used as the evaluation model, we found that the Bencher's performance degrades drastically by 25 points when comparing system pairs whose ChatBot Arena ratings differ by approximately 40 points, which is similar to the difference between qwen-1.5-72B and gpt-4-0314.

\subsection{RQ3: Can we use instance-level rankings of evaluation models as a good reference to select evaluation models for LLM benchers?}

\begin{figure}[h]
\centering
  \includegraphics[width=0.96\linewidth]{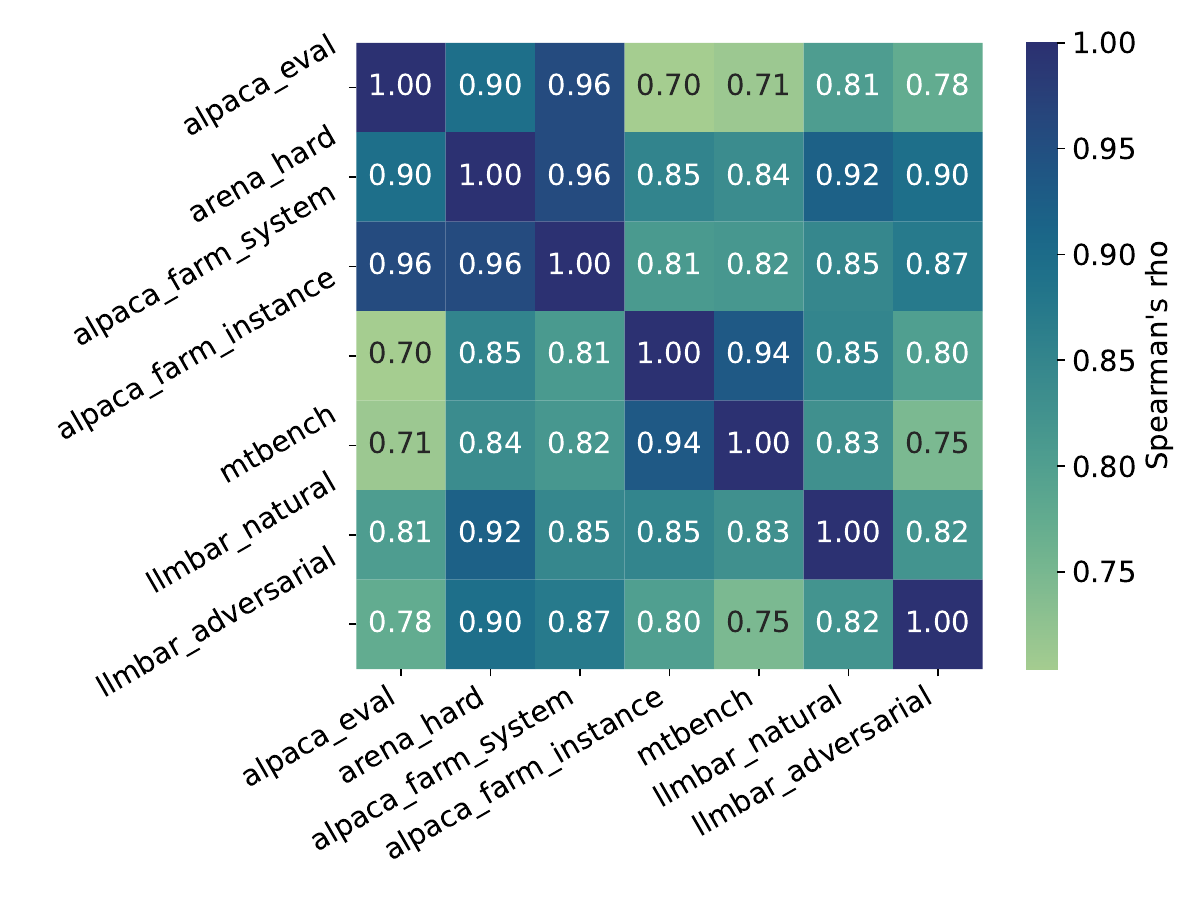}
  \caption{Spearman's $\rho$ between the evaluation model rankings under different meta-evaluation settings. \textbf{alpaca\_eval} and \textbf{arena\_hard} denote the evaluation model rankings are produced under \textbf{Setting 1}. \textbf{alpaca\_farm\_system} denotes that evaluation model rankings are produced under \textbf{Setting 3} where the instance-level human judgment of Alpaca Farm is used as ground truth, the Bradley-Terry model is used to aggregate both automatic evaluation and human evaluation. \textbf{alpaca\_farm\_instance}, \textbf{mtbench}, \textbf{llmbar\_natural}, and \textbf{llmbar\_adversarial} refer to that the evaluation model rankings are produced under \textbf{Setting 2} where instance-level human judgment of Alpaca Farm, MT-Bench, LLMBAR are used.}
\label{fig:heatmap_metaeval}
\end{figure}

Figure \ref{fig:heatmap_metaeval} displays the Spearman’s $\rho$ between evaluation model rankings obtained from different datasets and settings. We observed the following key findings: First, the evaluation model rankings vary considerably across different settings, as evidenced by the fact that none of the Spearman’s $\rho$ values are found to be 1.0. Second, system-level evaluation results are more consistent with each other. The evaluation model ranking obtained from Setting 3 is very similar to that from Setting 1, despite the systems to be evaluated and input sets in Alpaca Farm being different from those in Setting 1. In contrast, although the input set and involved systems in \texttt{alpaca\_farm\_instance} (Setting 2) and \texttt{alpaca\_farm\_system} (Setting 3) are exact the same, the evaluation model rankings they produce differ substantially. 
This suggests that whether the aggregation methods are applied has a great impact on the evaluation results. In addition, compared to \texttt{llmbar\_adversarial}, the evaluation model ranking from \texttt{llmbar\_natural} is more closely aligned with that from the system-level evaluation results. This is probably due to the responses in \texttt{llmbar\_adversarial} are not from real systems, which makes it more different from system-level evaluation setting.

\section{Related Work}

\textbf{Automatic LLM benchers.} 
This kind of research always create a automatic leaderboard for various LLMs. As mentioned above, the main contribution of Arena Hard \citep{DBLP:journals/corr/abs-2406-11939} lies in designing a pipeline for automatically constructing a challenging input set, also for Wildbench \citep{DBLP:journals/corr/abs-2406-04770}. In contrast, Alpaca Eval 2 \citep{DBLP:journals/corr/abs-2404-04475}, focus on designing better evaluation models and use existing instruction datasets as the input set. Besides, MixEval \citep{DBLP:journals/corr/abs-2406-06565} build LLM benchmarks by strategically mixing off-the-shelf ground-truth-based benchmarks that match real-world user queries. \citet{DBLP:journals/corr/abs-2405-20267} automate the entire LLM evaluation process using LLM-powered agents. There is another line of research that specifically focuses on how build LLM-based evaluation models \citep{DBLP:journals/corr/abs-2402-01383}.

\textbf{Comparison between different evaluation types and aggregation methods.}
\citet{DBLP:conf/eacl/LiusieMG24} compared the effectiveness of LLM-based evaluation models with absolute score prediction and pairwise comparison on several NLG tasks. 
\citet{liu2024reifereevaluatinginstructionfollowingevaluation} investigated which evaluation type is better under the instance-level evaluation settings of instruction following thoroughly. However, these studies are not oriented to system-level evaluation and aggregation methods are not involved. \citet{DBLP:journals/corr/abs-2403-16950} proposed a more efficient rank aggregation method, but it is also used for ranking different instances rather than systems.
\citet{DBLP:conf/acl/PeyrardZE020} showed the advantages of the Bradley-Terry model over mean and median on aggregating instance-level scores of NLP systems but this study focuses on what aggregation method makes more sense to choose as part of an evaluation metric.  \citet{daynauth2024rankingunraveledrecipesllm} proposed three desirable properties of aggregation methods and evaluated the robustness of several aggregation methods. It is worth mentioning that a concurrent work also focused on automatic system-level LLM judges and analyzed their decisiveness and bias \citep{gera2024justrankbenchmarkingllmjudges}.

\textbf{Instance-level evaluation of evaluation models.} A lot of studies have focused on collecting instance-level human judgments to evaluate LLM-based evaluation models \citep{DBLP:conf/iclr/ZengYG0G024,DBLP:conf/nips/ZhengC00WZL0LXZ23,DBLP:conf/nips/DuboisLTZGBGLH23,DBLP:journals/corr/abs-2308-01862,DBLP:conf/iclr/WangYYZYW0J000024,DBLP:conf/iclr/LiSYF0024,DBLP:conf/iclr/YeKKHKJTKS24}. However, to the best of our knowledge, no work has examined whether the evaluation results of them are consistent with the system-level settings.

\section{Conclusions}
We have explored how to construct effective LLM benchers and evaluate them. We provide several recommendations on how to combine the four components of an LLM bencher, analyze the current evaluation settings for LLM benchers and the evaluation models that serve as their components, and highlight the shortcomings in these settings. Our code and data will be available at \url{https://github.com/yale-nlp/RealRank}.

\section*{Limitations}
The observations and conclusions of this paper are dependent on the datasets and selected LLMs, and cannot be guaranteed to apply under all circumstances. Furthermore, the open-source models included as evaluation models are relatively small in scale, which may not reflect the evaluation capabilities of larger open-source models.


\bibliography{custom}

\appendix

\section{Conversion of 5-Point Pairwise and Pointwise Evaluation Type to Base Pairwise Type}
\label{appendix:conversion}

To convert 5-point pairwise comparisons into base pairwise comparisons, we apply the following rules as \citet{DBLP:journals/corr/abs-2406-11939}:
\begin{itemize}
    \item $o_{ij} \gg o_{kj}$ or $o_{ij} \ll o_{kj}$ converts to 6 instances of $o_{ij} > o_{kj}$ or $o_{ij} < o_{kj}$
    \item $o_{ij} > o_{kj}$ or $o_{ij} < o_{kj}$ converts to 2 instances of $o_{ij} > o_{kj}$ or $o_{ij} < o_{kj}$
    \item $o_{ij} = o_{kj}$ converts to 1 instance of $o_{ij} > o_{kj}$ and 1 instance of $o_{ij} < o_{kj}$
\end{itemize}

For pointwise evaluation type, the conversion follows this rule, and the total number of its instances is expanded to $mn(n-1)/2$.

{\small
\begin{align*}
& \mathcal{T}_{\text{base.pairwise}}(\mathcal{E}, x_j, o_{ij}, o_{kj}) = \\
& \begin{cases} 
o_{ij} > o_{kj} & \text{if } \mathcal{T}_{\text{pointwise}}(\mathcal{E}, x_j, o_{ij}) > \mathcal{T}_{\text{pointwise}}(\mathcal{E}, x_j, o_{kj}) \\
o_{ij} < o_{kj} & \text{if } \mathcal{T}_{\text{pointwise}}(\mathcal{E}, x_j, o_{ij}) < \mathcal{T}_{\text{pointwise}}(\mathcal{E}, x_j, o_{kj})
\end{cases}
\end{align*}
}

\section{Definitions of Spearman's \texorpdfstring{$\rho$}{rho} and Kendall's \texorpdfstring{$\tau$}{tau}}

\begin{align*}
\rho(R_A, R_H) = 1 - \frac{6 \sum_{i=1}^n (R_A(s_i) - R_H(s_i))^2}{n(n^2 - 1)}
\end{align*}

Where $R_A(s_i)$ is the rank of system $s_i$ produced by the automatic bencher and $n$ is the total number of systems being ranked.
\begin{align*}
\tau(R_A, R_H) = \frac{C - D}{\sqrt{(C + D + T_A)(C + D + T_H)}}
\end{align*}

Where $C$ is the number of concordant pairs (i.e., pairs of systems $(s_i, s_k)$ where the rank order between $s_i$ and $s_k$ is the same in both $R_A$ and $R_H$). $D$ is the number of discordant pairs (i.e., pairs where the rank order differs between $R_A$ and $R_H$). $T_A$ is the number of ties in the automatic bencher’s ranking $R_A$ (i.e., where two systems have the same rank). $T_H$ is the number of ties in the human-provided ranking $R_H$.

\section{Experimental Setup}
\label{appendix:experimental_setup}

\textbf{Input Set.} For RQ1 and RQ2, we selected two input sets, Arena Hard \citep{DBLP:journals/corr/abs-2406-11939} and Alpaca Eval \citep{alpaca_eval}, containing 500 and 805 inputs, respectively. 

\noindent \textbf{Systems.} 18 LLMs that ranked highly on Chatbot Arena were chosen as the systems to be evaluated\footnote{gpt-4-turbo-2024-04-09, claude-3-opus-20240229, claude-3-sonnet-20240229, gpt-4-0314, gpt-4-0613, mistral-large-2402, mistral-medium, qwen1.5-72b-chat, claude-2.0, claude-2.1, gpt-3.5-turbo-0613, mixtral-8x7b-instruct-v0.1, yi-34b-chat, gemini-pro, dbrx-instruct-preview, tulu-2-dpo-70b, vicuna-33b, starling-lm-7b-alpha}. The data from Chatbot Arena as of July 30, 2024, was used to compute the scores and rankings, serving as the ground truth. We reused the responses from the systems on Arena Hard, collected by \citet{DBLP:journals/corr/abs-2406-11939}, and the responses on Alpaca Eval, collected by \citet{alpaca_eval}.

\noindent \textbf{Evaluation Models.} We selected 10 LLMs, including four proprietary OpenAI models and six open-source models, to serve as evaluation models. These include models gpt-4-turbo (gpt-4-turbo-2024-04-09 \citep{DBLP:journals/corr/abs-2303-08774}), gpt-4o (gpt-4o-2024-08-06), gpt-4o-mini (gpt-4o-mini-2024-07-18), gpt-3.5-turbo (gpt-3.5-turbo-0125), llama-3.1-70b (\href{https://huggingface.co/meta-llama/Llama-3.1-70B}{meta-llama/Llama-3.1-70B} \citep{dubey2024llama}), qwen-2.5-72b (\href{https://huggingface.co/Qwen/Qwen2.5-72B-Instruct}{Qwen/Qwen2.5-72B-Instruct} \citep{yang2024qwen2}), mixtral-8x7b (\href{https://huggingface.co/mistralai/Mixtral-8x7B-Instruct-v0.1}{mistralai/Mixtral-8x7B-Instruct-v0.1}), llama-3.1-8b (\href{https://huggingface.co/meta-llama/Llama-3.1-8B}{meta-llama/Llama-3.1-8B}), glm-4-9b (\href{https://huggingface.co/THUDM/glm-4-9b-chat}{THUDM/glm-4-9b-chat} \citep{DBLP:journals/corr/abs-2406-12793}), mistral-7b-v0.3 (\href{https://huggingface.co/mistralai/Mistral-7B-Instruct-v0.3}{mistralai/Mistral-7B-Instruct-v0.3}), mistral-7b-v0.1 (\href{https://huggingface.co/mistralai/Mistral-7B-Instruct-v0.1}{mistralai/Mistral-7B-Instruct-v0.1} \citep{DBLP:journals/corr/abs-2310-06825}), llama-2-7b (\href{https://huggingface.co/meta-llama/Llama-2-7b-chat-hf}{meta-llama/Llama-2-7b-chat-hf}).

\noindent \textbf{Evaluation Types.} The evaluation types were implemented as follows. As for pointwise, given an input and a response, evaluation models were asked to score the response on a 0-9 scale. We followed the method of \citet{DBLP:conf/emnlp/LiuIXWXZ23}, using token probability to compute a weighted sum of the model's output for the 0-9 scale as the final instance-level score. For base Pairwise or 5-Point Pairwise, given an input and two responses, evaluation models were asked to either choose the better response or make a five-category comparison. The results of the automatic evaluations were extracted using regular expressions. If the evaluation model did not provide the requested judgment, a random evaluation result was assigned to the instance. The specific prompt designs can be found in the appendix.

\noindent \textbf{Aggregation Methods.} We adopted Chatbot Arena's implementation of the Bradley-Terry model and implemented other aggregation methods ourselves.

\noindent \textbf{Additional Setup for RQ3.} We selected three datasets containing instance-level human judgments: Alpaca Farm \citep{DBLP:conf/nips/DuboisLTZGBGLH23}, LLMBAR \citep{DBLP:conf/iclr/ZengYG0G024}, and MT-bench \citep{DBLP:conf/nips/ZhengC00WZL0LXZ23}. Among these, Alpaca Farm has a larger data scale, and 22 systems are involved, making it suitable for both Setting 2 and Setting 3. The number of systems in MT-bench and LLMBAR is too small, or system information is unavailable, so they can only be used for Setting 2. Additionally, LLMBAR contains two subsets: natural and adversarial. The natural subset contains responses from real systems, while the responses of the adversarial subset were generated using adversarial techniques.

\section{Discussions on Reference Systems}
\label{appendix:reference_system}
\begin{figure*}[]
  \includegraphics[width=\linewidth]{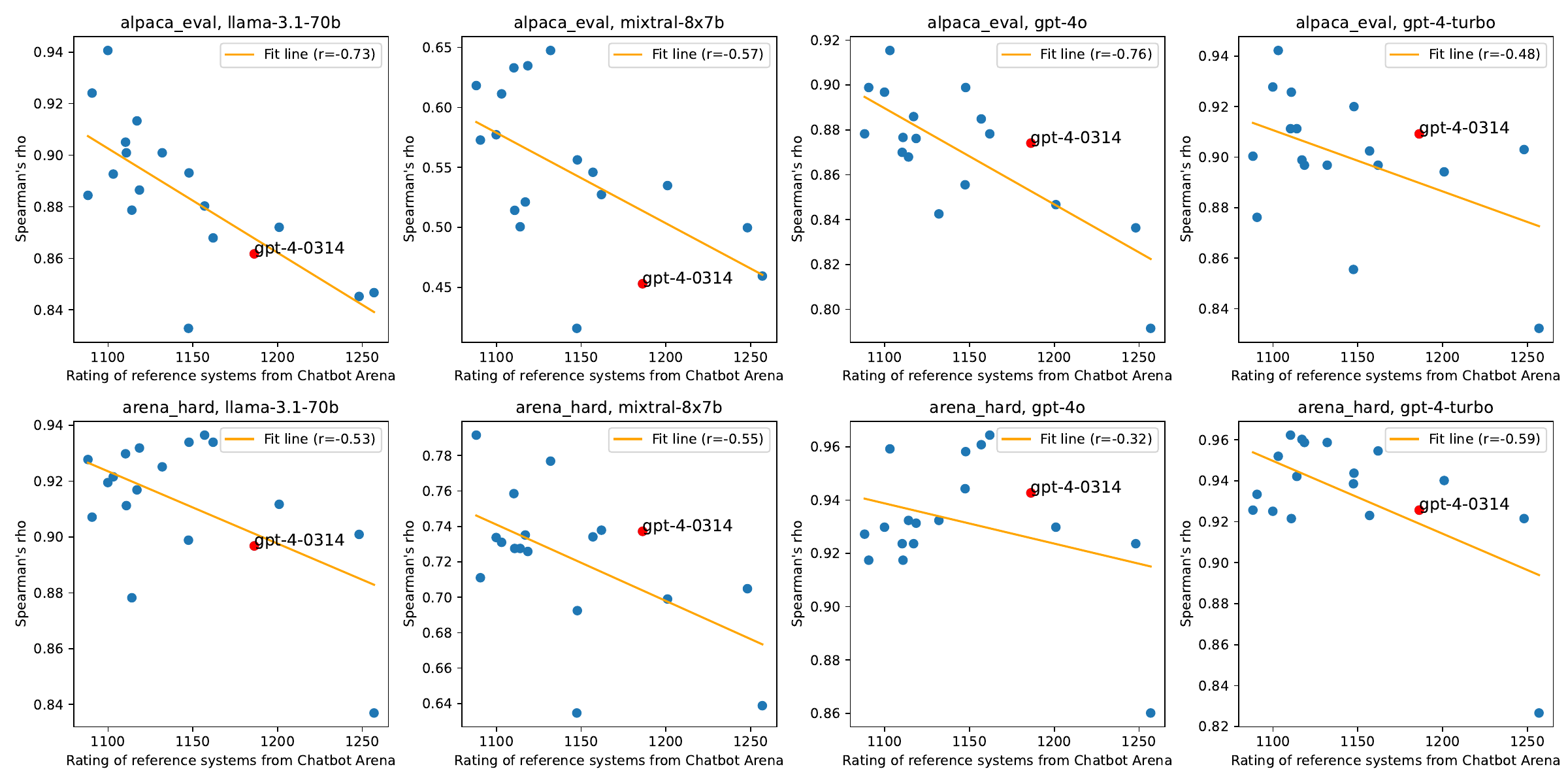}
  \caption{The relationships between the performance of the bencher and the Chatbot Arena ratings of the LLM used as the reference systems. The evaluation type is fixed as base pairwise.}
  \label{fig:reference_system_rating1}
\end{figure*}

\begin{figure*}[]
  \includegraphics[width=\linewidth]{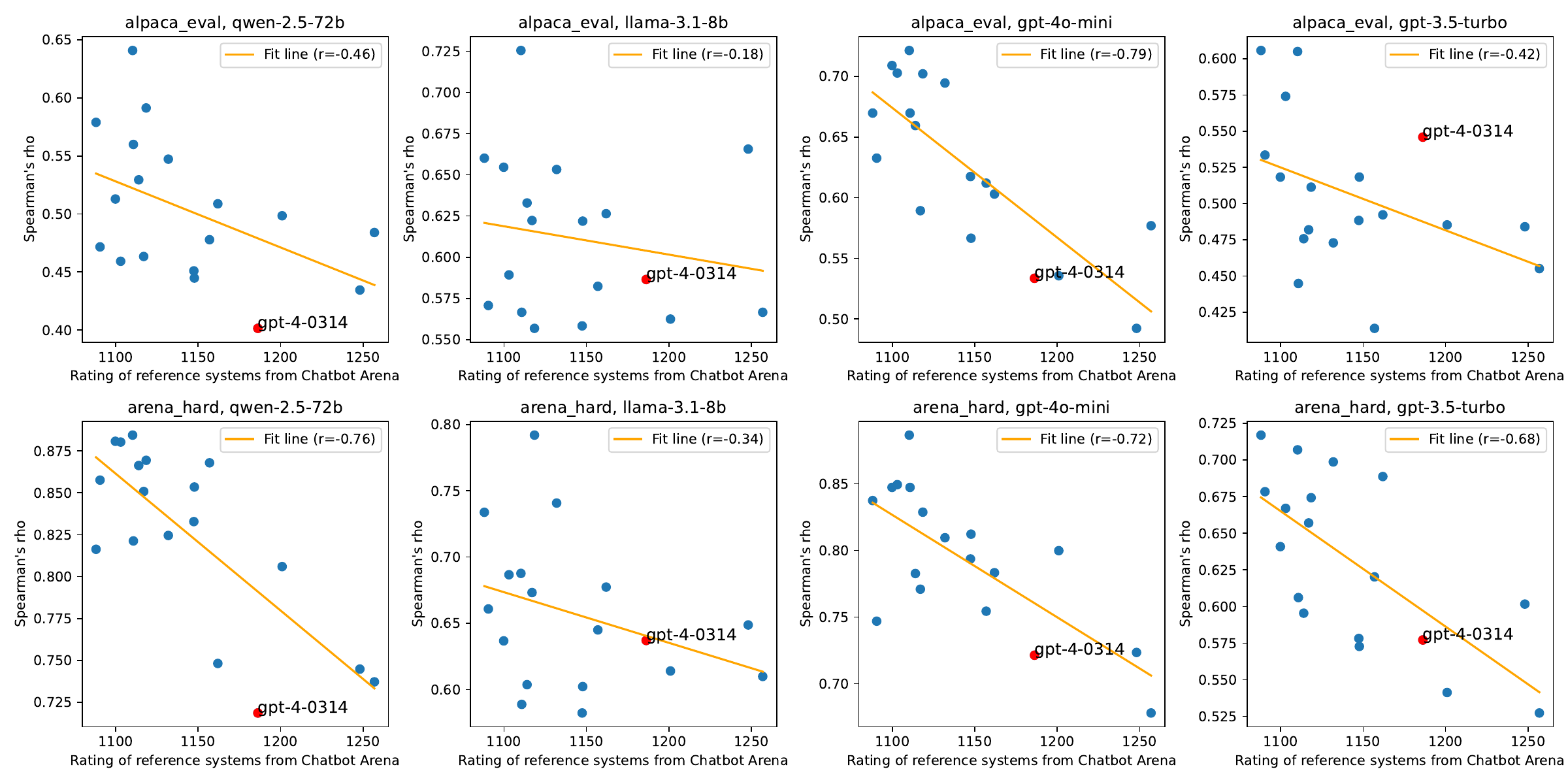}
  \caption{The relationships between the performance of the bencher and the Chatbot Arena ratings of the LLM used as the reference systems. The evaluation type is fixed as base pairwise.}
  \label{fig:reference_system_rating2}
\end{figure*}

\begin{figure*}[]
  \includegraphics[width=\linewidth]{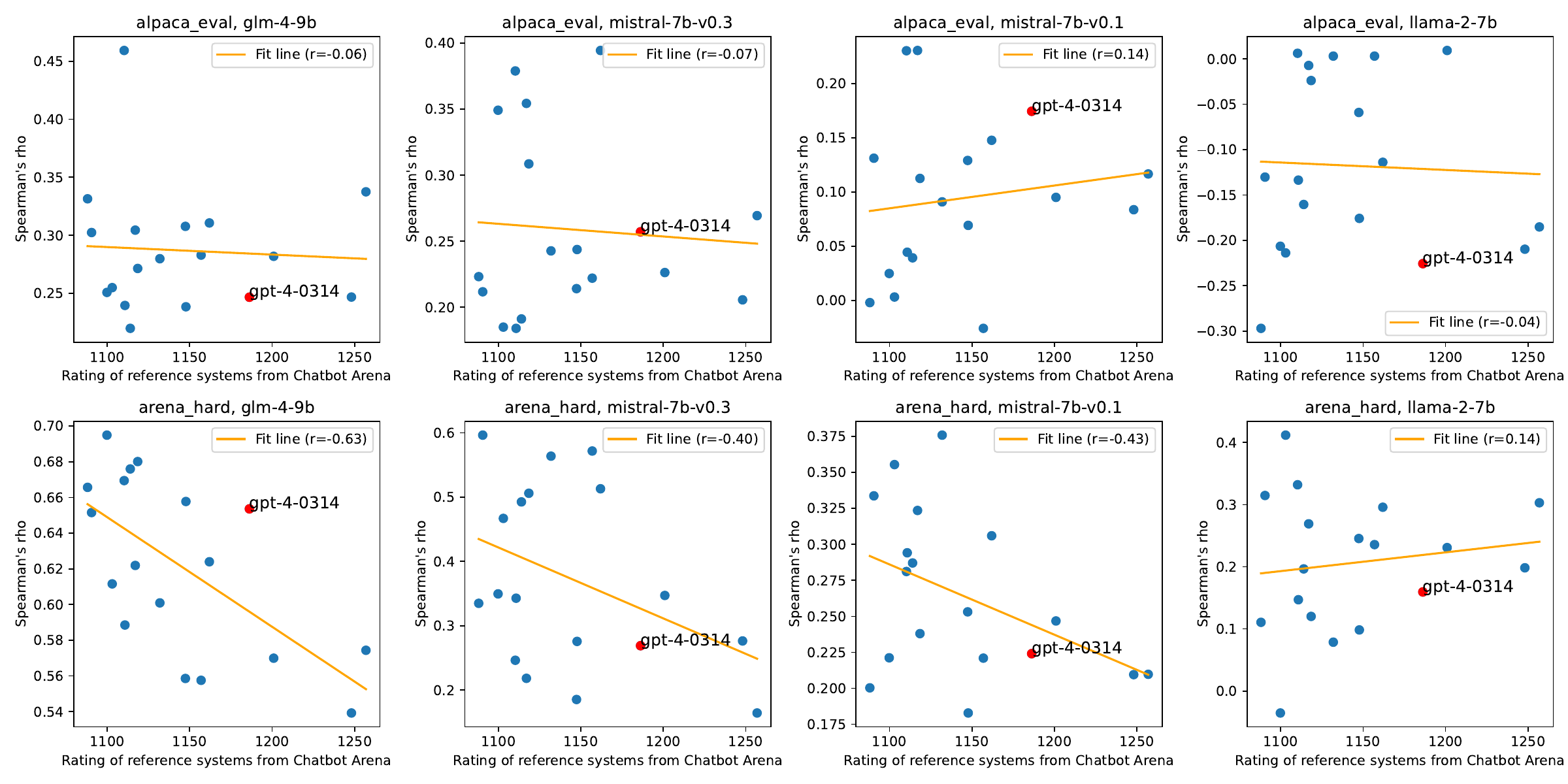}
  \caption{The relationships between the performance of the bencher and the Chatbot Arena ratings of the LLM used as the reference systems. The evaluation type is fixed as base pairwise.}
  \label{fig:reference_system_rating3}
\end{figure*}

Although using a strong system as a reference system is a common practice, its rationale has not been thoroughly examined. Therefore, we calculated the performance of the bencher when each of the 18 evaluated systems is used as the reference system and demonstrated the relationship between the bencher’s performance and the Chatbot Arena rating of the LLM used as the reference system, as shown in Figures \ref{fig:reference_system_rating1}, \ref{fig:reference_system_rating2}, and \ref{fig:reference_system_rating3}. We found that, for most strong evaluation models, the performance of the bencher is negatively correlated with the level of the LLM used as the reference system. Therefore, we believe this issue warrants further investigation, and at the very least, we should be more cautious in selecting the reference system.

\section{Cost Analysis}
\label{appendix:cost_analysis}

To perform a more granular cost analysis of the bencher, we measure cost in terms of the total number of tokens consumed during automatic evaluation and the evaluation model's cost per million tokens. For OpenAI models, we use the pricing for the Batch API\footnote{\url{https://openai.com/api/pricing/}} provided on their official website. For open-source models, we estimate costs based on electricity consumption during inference on GPU servers, ignoring the expenses of purchasing and maintaining GPUs.

Specifically, we estimate the number of tokens processed per second by open-source models based on throughput data from the vLLM blog\footnote{\url{https://blog.vllm.ai/2024/09/05/perf-update.html}}:

\begin{itemize}
    \item \textbf{8B Model on 1xH100:} Approximately 30 queries per second (QPS) with an average of 179 output tokens per query. Tokens per second: $30 \times 179 = 5370 \, \text{tokens/second}$.
    
    \item \textbf{70B Model on 4xH100:} Approximately 15 QPS with an average of 179 output tokens per query. Tokens per second: $15 \times 179 = 2685 \, \text{tokens/second}$.
\end{itemize}

Next, we calculate the time required to process 1 million tokens:

\begin{itemize}
    \item $\textbf{8B Model:} \frac{1000000}{5370} \approx 186.2 \, \text{seconds}.$
    \item $\textbf{70B Model:} \frac{1000000}{2685} \approx 372.4 \, \text{seconds}.$
\end{itemize}

We then estimate the energy consumption for processing 1 million tokens:

\begin{itemize}
    \item \textbf{8B Model on 1xH100:}
    $
    \text{Energy consumed} = \text{Power (700W)} \times \text{Time (186.2 seconds)} = 130340 \, \text{watt-seconds} \approx 36.21 \, \text{watt-hours (Wh)}.
    $

    \item \textbf{70B Model on 4xH100:}
    $
    \text{Energy consumed per GPU} = \text{Power (700W)} \times \text{Time (372.4 seconds)} = 260680 \, \text{watt-seconds} \approx 72.41 \, \text{Wh}, \\
    \text{Total energy for 4 GPUs} = 72.41 \, \text{Wh} \times 4 = 289.6 \, \text{Wh}.
    $
\end{itemize}

Finally, assuming an electricity rate of \$0.17 per kilowatt-hour (kWh), which is the average rate in the US\footnote{\url{https://www.bls.gov/regions/midwest/data/averageenergyprices_selectedareas_table.htm}}:

\begin{itemize}
    \item \textbf{8B Model:}
    $
    \text{Cost} = 36.21 \, \text{Wh} \times \frac{\$0.17}{1000 \, \text{Wh}} = \$0.00577 \, (\sim\$0.006) \, \text{per 1M tokens}.
    $

    \item \textbf{70B Model:}
    $
    \text{Cost} = 289.6 \, \text{Wh} \times \frac{\$0.17}{1000 \, \text{Wh}} = \$0.0492 \, (\sim\$0.05) \, \text{per 1M tokens}.
    $
\end{itemize}

\begin{table}[]
\centering
\begin{tabular}{ll}
\toprule
Model & \$/1M tokens \\
\midrule
llama-3.1-8b & 0.006 \\
llama-3.1-70b & 0.05 \\
qwen-2.5-72b & 0.05 \\
gpt-4o & 1.25 \\
gpt-4-turbo & 5.00 \\
gpt-3.5-turbo & 0.25 \\
gpt-4o-mini & 0.075 \\
\bottomrule
\end{tabular}
\caption{Financial cost of different evaluation models during inference. The prices of open-source models are estimated by electricity rate. We use the cost of input tokens because our output length is negligible.}
\label{tab:model_cost_electricity}
\end{table}

\begin{figure*}[]
  \includegraphics[width=\linewidth]{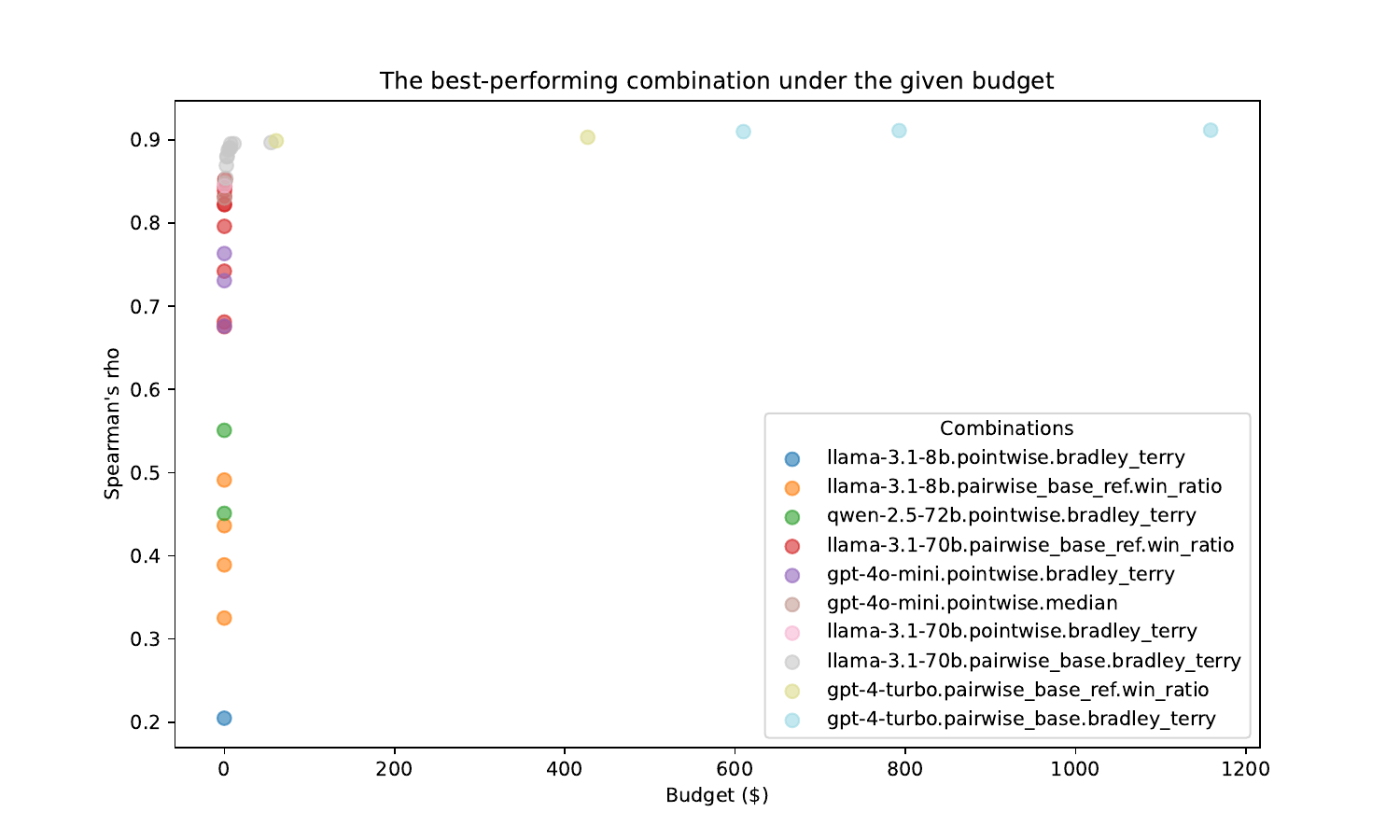}
  \caption{The best-performing combinations under varying budgets on Alpaca Eval.}
  \label{fig:cost_analysis_alpaca_eval}
\end{figure*}

\begin{figure*}[]
  \includegraphics[width=\linewidth]{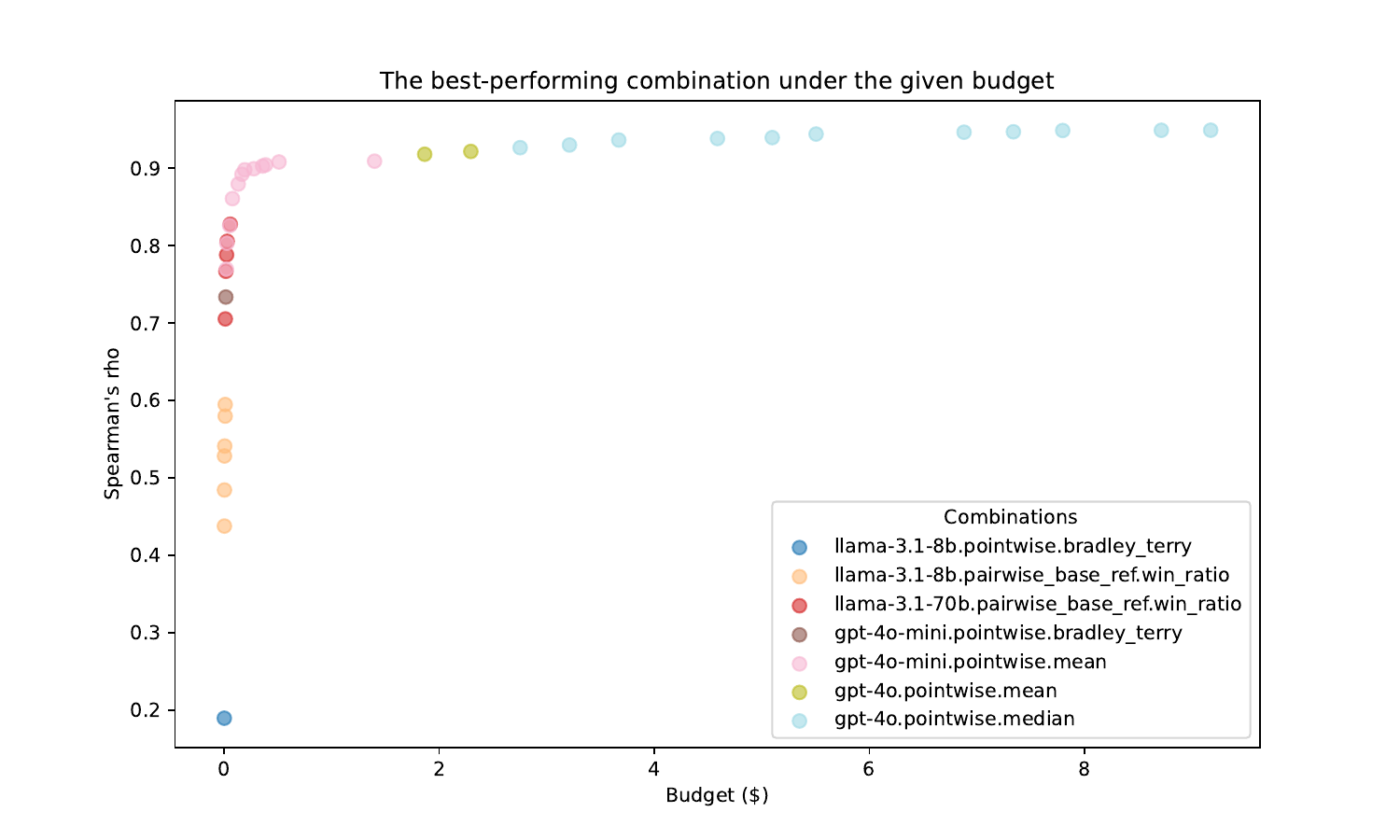}
  \caption{The best-performing combinations under varying budgets on Arena Hard. Using OpenAI models as evaluation models and employing pairwise evaluation incurs higher costs but does not outperform pointwise evaluation.}
  \label{fig:cost_analysis_arena_hard}
\end{figure*}

Thus, we derived the cost per 1M tokens for each model, as shown in Table \ref{tab:model_cost_electricity}. Furthermore, we performed sampling with replacement on the input set at proportions of 0.02, 0.03, 0.04, 0.05, 0.10, and up to 1.00 to generate input sets of varying sizes \footnote{Repeating 100 times and taking the average performance.}. Consequently, the cost of different combinations of evaluation models, evaluation types, and aggregation methods across input sets of different sizes can be estimated. Figures \ref{fig:cost_analysis_alpaca_eval} and \ref{fig:cost_analysis_arena_hard} illustrate the best-performing combinations under varying budget constraints. From these results, we observe that open-source models demonstrate a significant advantage when the budget is highly constrained.

\section{Confidence Interval Analysis}
\begin{figure*}[]
  \includegraphics[width=\linewidth]{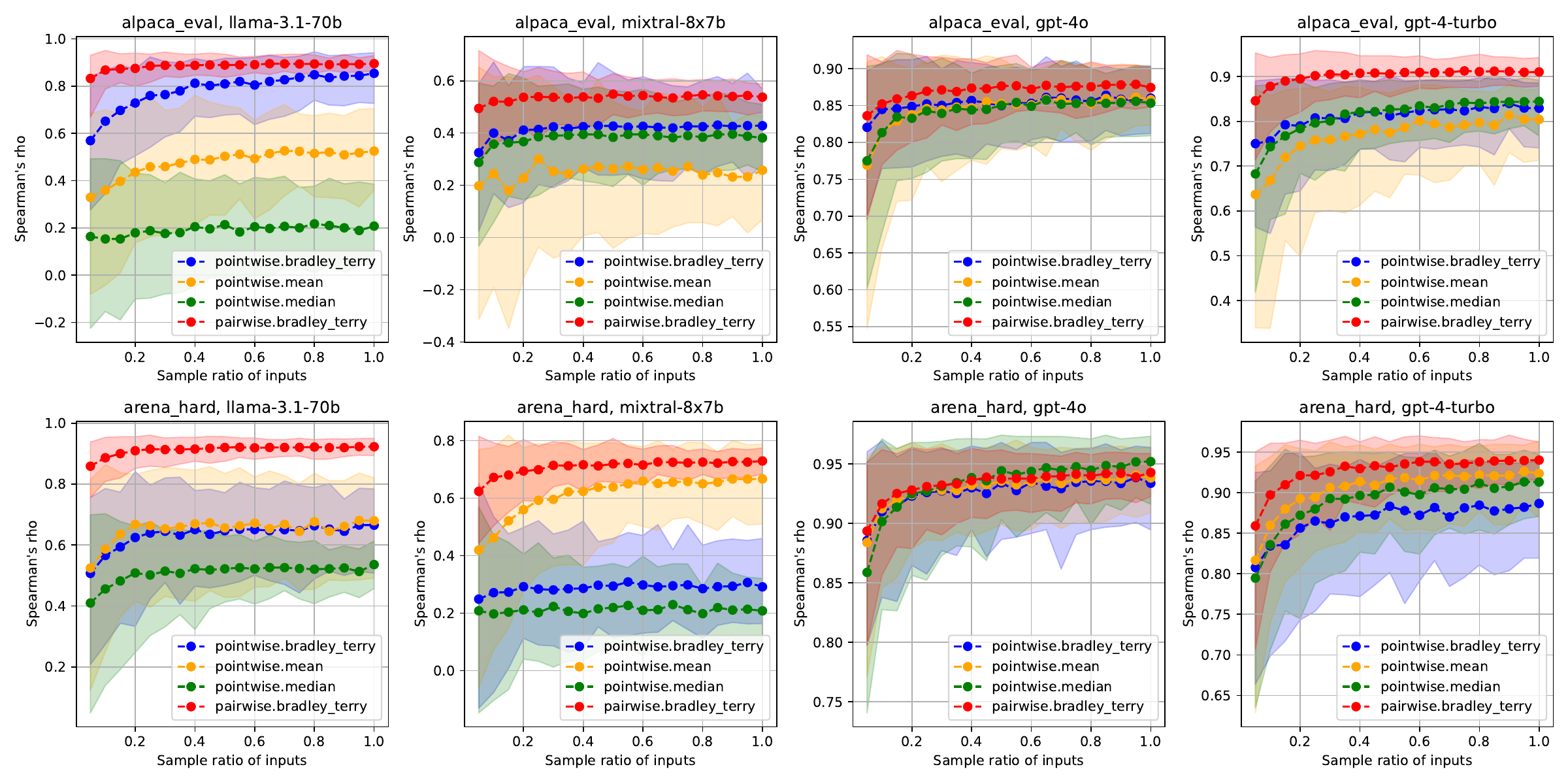}
  \caption{Bootstrapping mean and 95\% confidence interval of the correlations between the system rankings of automatic evaluation  metrics and human judgment. The input set is sampled with replacement. The number of iterations is set to 100.}
  \label{fig:input_sampling}
\end{figure*}

\begin{figure*}[]
  \includegraphics[width=\linewidth]{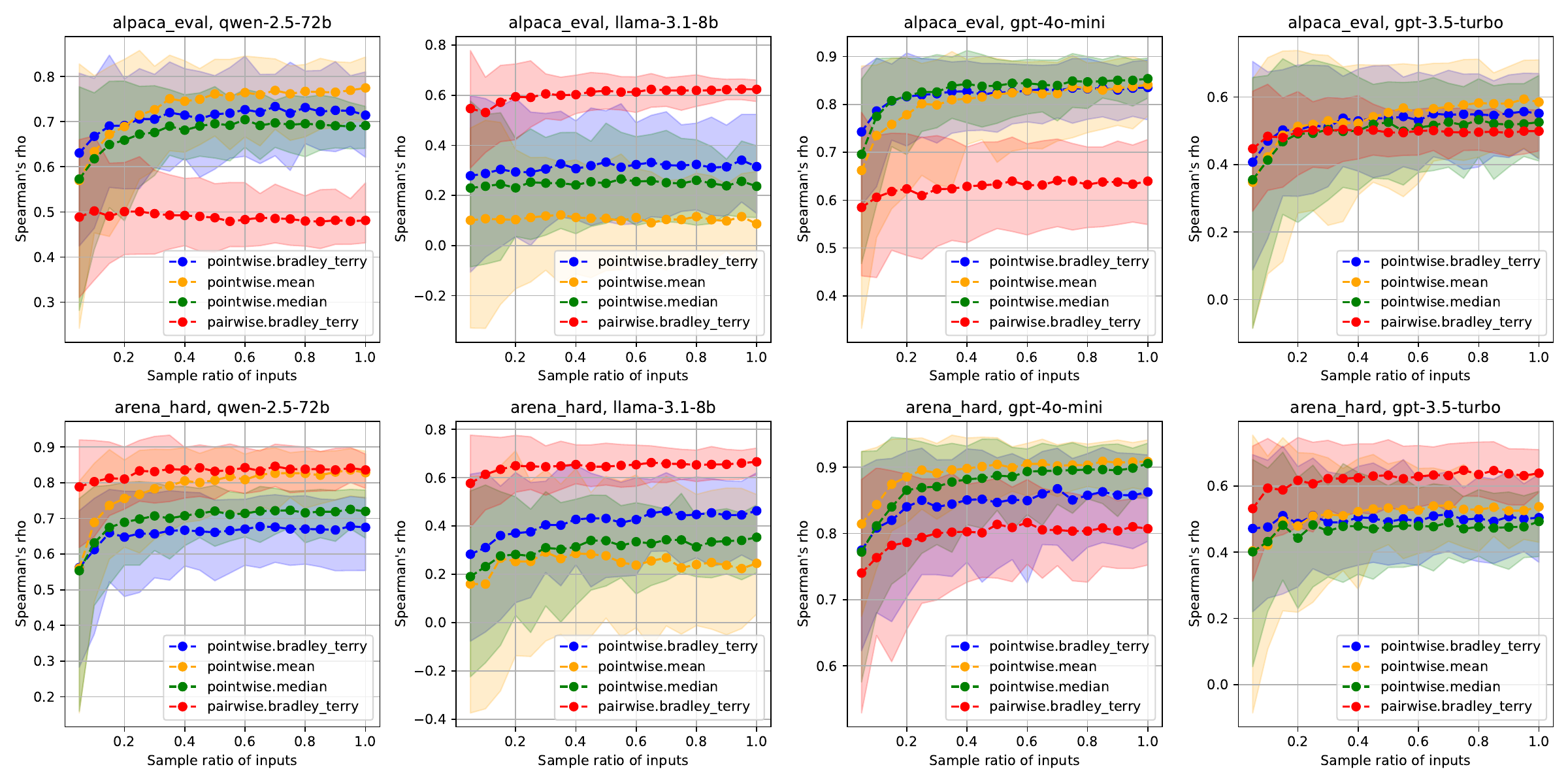}
  \caption{Bootstrapping mean and 95\% confidence interval of the correlations between the system rankings of automatic evaluation  metrics and human judgment. The input set is sampled with replacement. The number of iterations is set to 100.}
  \label{fig:input_sampling2}
\end{figure*}

\begin{figure*}[]
  \includegraphics[width=\linewidth]{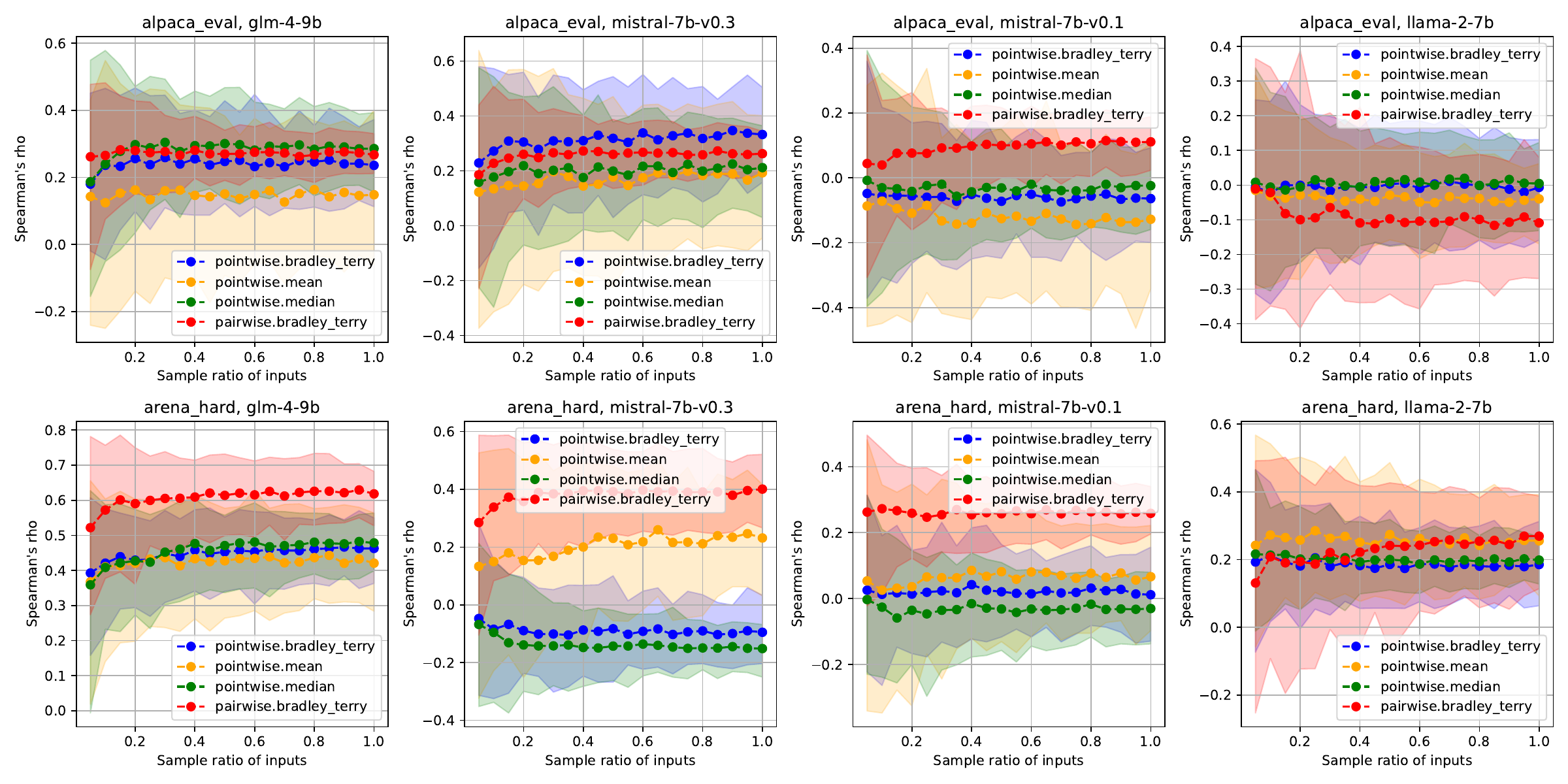}
  \caption{Bootstrapping mean and 95\% confidence interval of the correlations between the system rankings of automatic evaluation  metrics and human judgment. The input set is sampled with replacement. The number of iterations is set to 100.}
  \label{fig:input_sampling3}
\end{figure*}

In Table \ref{tab:type_aggregation_main}, we computed the performance of benchers of different combinations of $\mathcal{X}, \mathcal{E}, \mathcal{T}, \mathcal{G}$, but these are merely point estimates. To assess the accuracy of our estimates, we obtained the 95\% confidence intervals for bencher performance through bootstrapping. As shown in Figure \ref{fig:input_sampling}, \ref{fig:input_sampling2}, and \ref{fig:input_sampling3}, with an increasing number of inputs, the confidence intervals for most benchers become narrower. However, for some combinations of evaluation type and aggregation method, the confidence intervals remain relatively wide when the sample ratio equals 1.0, indicating a high degree of uncertainty in the performance estimates. In contrast, the combination of base pairwise and the Bradley-Terry model has narrower confidence intervals (the red section), suggesting that the performance estimates for this combination are more precise.

\section{Other Figures and Tables}

\begin{table*}[]
\small
\centering
\begin{tabular}{llcccccccc}
\toprule
\multicolumn{10}{c}{Alpaca Eval as the input set} \\ \midrule
\multicolumn{2}{c}{Evaluation Model} & \multicolumn{2}{c}{qwen-2.5-72b} & \multicolumn{2}{c}{llama-3.1-8b} & \multicolumn{2}{c}{gpt-4o-mini} & \multicolumn{2}{c}{gpt-3.5-turbo} \\ \midrule
Type & \multicolumn{1}{l}{Aggregation} & $\rho$ & $\tau$ & $\rho$ & $\tau$ & $\rho$ & $\tau$ & $\rho$ & $\tau$ \\ 
\midrule
pointwise & bradley\_terry & 0.8101 & 0.6471 & 0.2982 & 0.1503 & 0.8452 & 0.6863 & 0.5707 & 0.4510 \\
pointwise & win\_ratio & 0.8101 & 0.6471 & 0.2982 & 0.1503 & 0.8452 & 0.6863 & 0.5707 & 0.4510 \\
pointwise & mean & \textbf{0.8576} & \textbf{0.7124} & 0.0423 & 0.0719 & \textbf{0.8700} & \textbf{0.7124} & \textbf{0.6574} & \textbf{0.4902} \\
pointwise & median & 0.7895 & 0.6209 & 0.2239 & 0.0980 & 0.8493 & 0.6863 & 0.5315 & 0.4248 \\ \midrule
pairwise\_base & bradley\_terry & 0.4592 & 0.3203 & \textbf{0.6347} & \textbf{0.4902} & 0.6821 & 0.5033 & 0.4923 & 0.3464 \\
pairwise\_base & win\_ratio & 0.4592 & 0.3203 & \textbf{0.6347} & \textbf{0.4902} & 0.6821 & 0.5033 & 0.4923 & 0.3464 \\ \midrule
pairwise\_5point & bradley\_terry & 0.4221 & 0.3072 & 0.4241 & 0.2810 & 0.4799 & 0.3333 & 0.4510 & 0.3725 \\
pairwise\_5point & win\_ratio & 0.2611 & 0.1895 & 0.2033 & 0.1503 & 0.4551 & 0.3203 & 0.3354 & 0.2549 \\ \midrule
pairwise\_base\_ref & bradley\_terry & 0.4014 & 0.2680 & 0.5996 & 0.4510 & 0.5397 & 0.4248 & 0.5459 & 0.3987 \\
pairwise\_base\_ref & win\_ratio & 0.4014 & 0.2680 & 0.5865 & 0.4328 & 0.5335 & 0.3987 & 0.5459 & 0.3987 \\ \midrule
pairwise\_5point\_ref & bradley\_terry & 0.6140 & 0.4641 & 0.3808 & 0.2680 & 0.5191 & 0.3464 & 0.1930 & 0.1242 \\
pairwise\_5point\_ref & win\_ratio & 0.3395 & 0.2418 & 0.3457 & 0.2680 & 0.4324 & 0.3072 & 0.0815 & 0.0588 \\
\midrule
\multicolumn{10}{c}{Arena Hard as the input set} \\ \midrule
\multicolumn{2}{c}{Evaluation Model} & \multicolumn{2}{c}{qwen-2.5-72b} & \multicolumn{2}{c}{llama-3.1-8b} & \multicolumn{2}{c}{gpt-4o-mini} & \multicolumn{2}{c}{gpt-3.5-turbo} \\ \midrule
Type & \multicolumn{1}{l}{Aggregation} & $\rho$ & $\tau$ & $\rho$ & $\tau$ & $\rho$ & $\tau$ & $\rho$ & $\tau$ \\ 
\midrule
pointwise & bradley\_terry & 0.6739 & 0.4641 & 0.5232 & 0.4379 & 0.8824 & 0.7255 & 0.5212 & 0.3987 \\
pointwise & win\_ratio & 0.6739 & 0.4641 & 0.5232 & 0.4379 & 0.8824 & 0.7255 & 0.5212 & 0.3987 \\
pointwise & mean & 0.8658 & 0.6993 & 0.1744 & 0.0980 & \textbf{0.9174} & \textbf{0.7908} & 0.5480 & 0.3856 \\
pointwise & median & 0.7069 & 0.5294 & 0.3333 & 0.2941 & 0.9112 & 0.7778 & 0.4840 & 0.3464 \\ \midrule
pairwise\_base & bradley\_terry & 0.8411 & 0.6863 & \textbf{0.6656} & \textbf{0.5556} & 0.8101 & 0.6732 & 0.6636 & \textbf{0.5686} \\
pairwise\_base & win\_ratio & 0.8411 & 0.6863 & \textbf{0.6656} & \textbf{0.5556} & 0.8101 & 0.6732 & 0.6636 & \textbf{0.5686} \\ \midrule
pairwise\_5point & bradley\_terry & \textbf{0.9236} & \textbf{0.7647} & 0.3086 & 0.2288 & 0.6491 & 0.4771 & \textbf{0.6656} & 0.5033 \\
pairwise\_5point & win\_ratio & 0.8989 & 0.7386 & 0.2797 & 0.2157 & 0.6491 & 0.4771 & 0.6120 & 0.4510 \\ \midrule
pairwise\_base\_ref & bradley\_terry & 0.7207 & 0.5771 & 0.6267 & 0.5246 & 0.7152 & 0.5817 & 0.5772 & 0.4721 \\
pairwise\_base\_ref & win\_ratio & 0.7186 & 0.5639 & 0.6371 & 0.5377 & 0.7214 & 0.5817 & 0.5772 & 0.4721 \\ \midrule
pairwise\_5point\_ref & bradley\_terry & 0.9030 & 0.7386 & 0.2549 & 0.1895 & 0.6347 & 0.4771 & 0.6161 & 0.4641 \\
pairwise\_5point\_ref & win\_ratio & 0.6801 & 0.5033 & 0.1786 & 0.1311 & 0.6285 & 0.4641 & 0.5872 & 0.4379 \\
\midrule
\end{tabular}
\caption{Standard meta-evaluation results of different combinations of input sets, evaluation models, evaluation types, and aggregation methods (the correlation between the automatic benchers' and ChatBot Arena's evaluations). For the pairwise evaluation using a reference system, we show the results with gpt-4-0314 as the reference system.}
\label{tab:type_aggregation_main-2}
\end{table*}

\begin{table*}[]
\small
\centering
\begin{tabular}{llcccccccc}
\toprule
\multicolumn{10}{c}{Alpaca Eval as the input set} \\ \midrule
\multicolumn{2}{c}{Evaluation Model} & \multicolumn{2}{c}{glm-4-9b} & \multicolumn{2}{c}{mistral-7b-v0.3} & \multicolumn{2}{c}{mistral-7b-v0.1} & \multicolumn{2}{c}{llama-2-7b} \\ \midrule
Type & \multicolumn{1}{l}{Aggregation} & $\rho$ & $\tau$ & $\rho$ & $\tau$ & $\rho$ & $\tau$ & $\rho$ & $\tau$ \\ 
\midrule
pointwise & bradley\_terry & 0.2693 & 0.2026 & 0.3354 & \textbf{0.2288} & -0.0898 & -0.0327 & -0.0072 & 0.0196 \\
pointwise & win\_ratio & 0.2693 & 0.2026 & 0.3354 & \textbf{0.2288} & -0.0898 & -0.0327 & -0.0072 & 0.0196 \\
pointwise & mean & 0.1063 & 0.0588 & 0.2219 & 0.1373 & -0.1930 & -0.1373 & -0.0506 & -0.0065 \\
pointwise & median & 0.2900 & \textbf{0.2157} & 0.2755 & 0.1895 & -0.0093 & 0.0327 & -0.0114 & 0.0065 \\ \midrule
pairwise\_base & bradley\_terry & 0.2755 & 0.1765 & 0.2549 & 0.1765 & 0.1476 & 0.1111 & -0.1291 & -0.0724 \\
pairwise\_base & win\_ratio & 0.2755 & 0.1765 & 0.2549 & 0.1765 & 0.1476 & 0.1111 & -0.1291 & -0.0724 \\ \midrule
pairwise\_5point & bradley\_terry & 0.2425 & 0.1503 & -0.0279 & 0.0196 & -0.1414 & -0.085 & \textbf{0.1269} & \textbf{0.0719} \\
pairwise\_5point & win\_ratio & 0.1765 & 0.1111 & -0.1517 & -0.0588 & -0.1414 & -0.085 & \textbf{0.1311} & \textbf{0.0588} \\ \midrule
pairwise\_base\_ref & bradley\_terry & \textbf{0.3065} & \textbf{0.2157} & \textbf{0.3397} & 0.2098 & \textbf{0.2095} & \textbf{0.1503} & -0.2183 & -0.1258 \\
pairwise\_base\_ref & win\_ratio & 0.2466 & 0.1765 & 0.2571 & 0.1574 & 0.1744 & 0.1242 & -0.2257 & -0.1329 \\ \midrule
pairwise\_5point\_ref & bradley\_terry & 0.2776 & 0.1634 & -0.1703 & -0.1242 & -0.1176 & -0.0821 & 0.0795 & 0.0588 \\
pairwise\_5point\_ref & win\_ratio & 0.2095 & 0.1242 & -0.2611 & -0.1765 & -0.0836 & -0.0525 & 0.0609 & 0.0588 \\
\midrule
\multicolumn{10}{c}{Arena Hard as the input set} \\ \midrule
\multicolumn{2}{c}{Evaluation Model} & \multicolumn{2}{c}{glm-4-9b} & \multicolumn{2}{c}{mistral-7b-v0.3} & \multicolumn{2}{c}{mistral-7b-v0.1} & \multicolumn{2}{c}{llama-2-7b} \\ \midrule
Type & \multicolumn{1}{l}{Aggregation} & $\rho$ & $\tau$ & $\rho$ & $\tau$ & $\rho$ & $\tau$ & $\rho$ & $\tau$ \\ 
\midrule
pointwise & bradley\_terry & 0.4696 & 0.3464 & -0.0918 & -0.0196 & 0.0237 & 0.0458 & 0.1723 & 0.0980 \\
pointwise & win\_ratio & 0.4696 & 0.3464 & -0.0918 & -0.0196 & 0.0237 & 0.0458 & 0.1723 & 0.0980 \\
pointwise & mean & 0.4345 & 0.3203 & 0.2755 & 0.1765 & 0.0918 & 0.0327 & \textbf{0.2879} & \textbf{0.1895} \\
pointwise & median & 0.5005 & 0.3595 & -0.1662 & -0.0719 & -0.0299 & 0.0065 & 0.2054 & 0.1373 \\ \midrule
pairwise\_base & bradley\_terry & 0.6367 & 0.4902 & \textbf{0.4076} & \textbf{0.2941} & \textbf{0.2755} & 0.1765 & 0.2839 & 0.1836 \\
pairwise\_base & win\_ratio & 0.6367 & 0.4902 & \textbf{0.4076} & \textbf{0.2941} & \textbf{0.2755} & 0.1765 & 0.2839 & 0.1836 \\ \midrule
pairwise\_5point & bradley\_terry & 0.5315 & 0.4118 & 0.1538 & 0.0850 & 0.2487 & 0.1503 & -0.1930 & -0.1634 \\
pairwise\_5point & win\_ratio & 0.4964 & 0.3725 & 0.2012 & 0.1242 & 0.2487 & 0.1503 & -0.1414 & -0.0980 \\ \midrule
pairwise\_base\_ref & bradley\_terry & \textbf{0.6536} & \textbf{0.5115} & 0.2690 & 0.1728 & 0.2654 & \textbf{0.1967} & 0.1998 & 0.1595 \\
pairwise\_base\_ref & win\_ratio & \textbf{0.6536} & \textbf{0.5115} & 0.2690 & 0.1728 & 0.2241 & 0.1705 & 0.1595 & 0.1267 \\ \midrule
pairwise\_5point\_ref & bradley\_terry & 0.5418 & 0.4248 & 0.2136 & 0.1242 & 0.0974 & 0.0490 & -0.4200 & -0.3203 \\
pairwise\_5point\_ref & win\_ratio & 0.5624 & 0.4510 & 0.1765 & 0.0850 & 0.2239 & 0.1503 & -0.4407 & -0.3464 \\
\midrule
\end{tabular}
\caption{Standard meta-evaluation results of different combinations of input sets, evaluation models, evaluation types, and aggregation methods (the correlation between the automatic benchers' and ChatBot Arena's evaluations). For the pairwise evaluation using a reference system, we show the results with gpt-4-0314 as the reference system.}
\label{tab:type_aggregation_main-3}
\end{table*}

\begin{figure*}[]
  \includegraphics[width=0.95\linewidth]{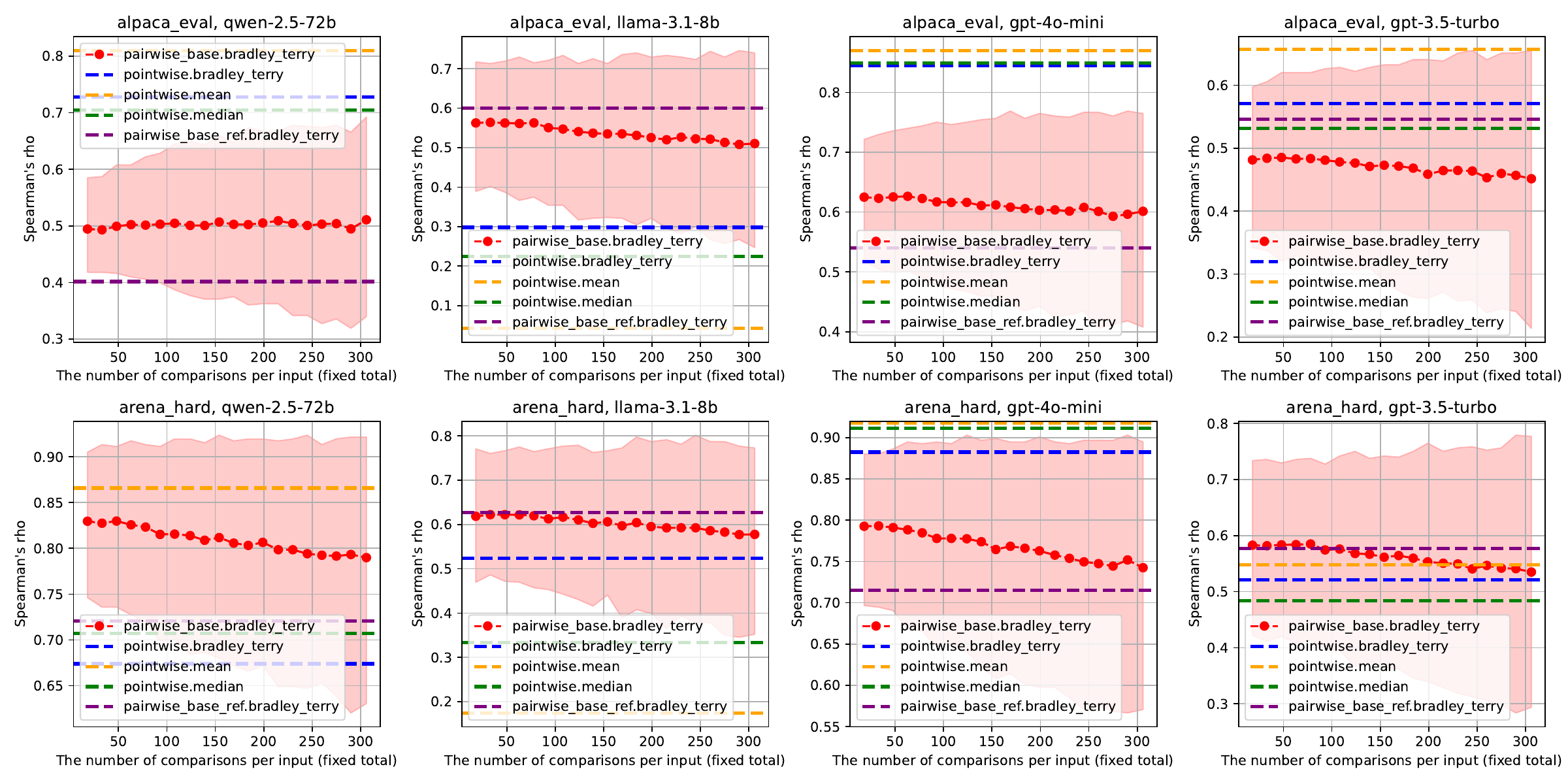}
  \caption{Bootstrapping mean and 95\% confidence interval of the correlations between the system rankings of automatic pairwise evaluation and human judgment when the total number of examples is set equal to that of pointwise evaluation. Results for pointwise and pairwise evaluations using gpt-4-0314 as the reference system are shown as horizontal lines. Both input sets and comparisons are sampled with replacement, with 1000 iterations.}
    \label{fig:p_pairs_sampling2}
\end{figure*}

\begin{figure*}[]
  \includegraphics[width=0.95\linewidth]{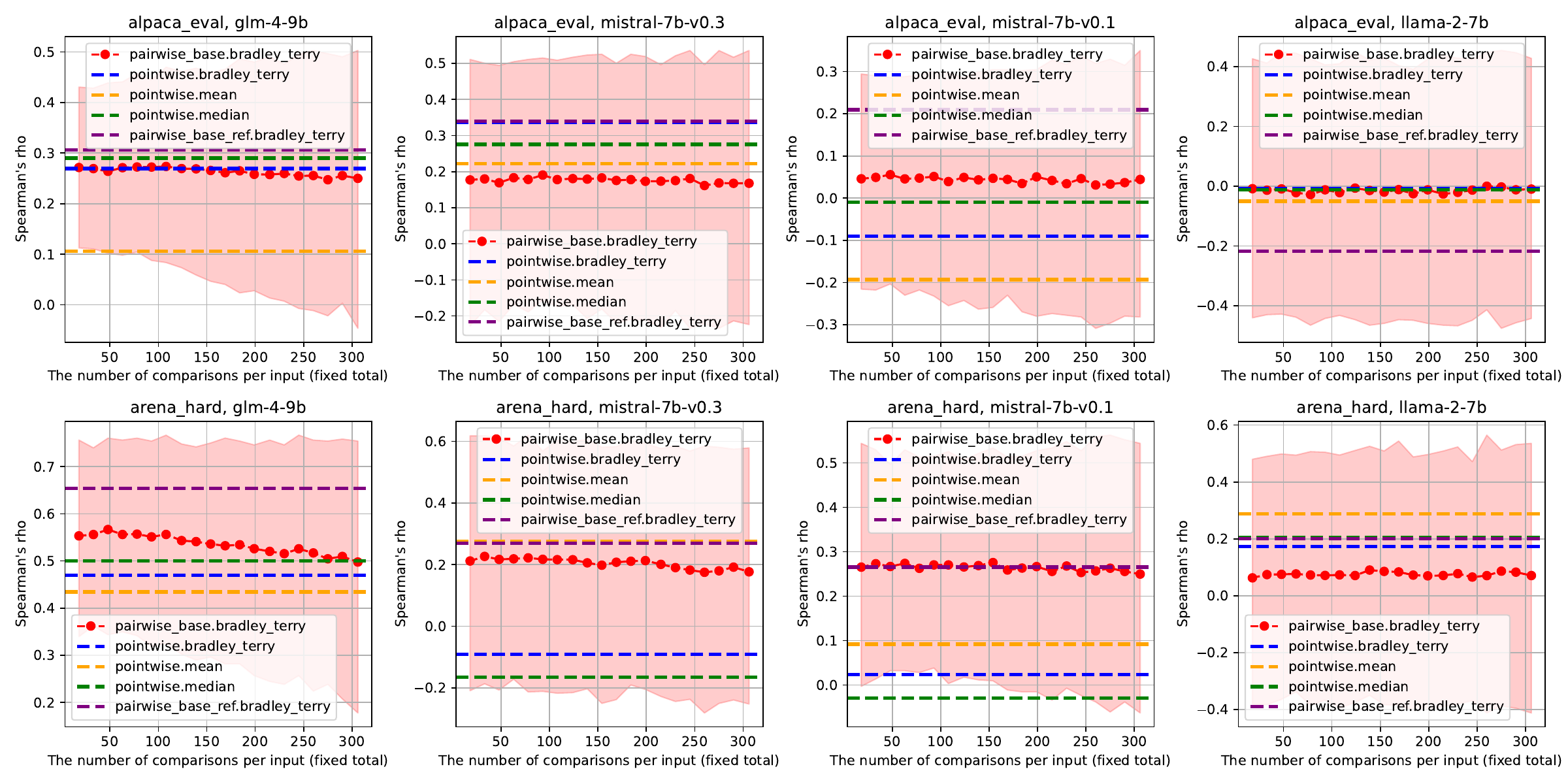}
  \caption{Bootstrapping mean and 95\% confidence interval of the correlations between the system rankings of automatic pairwise evaluation and human judgment when the total number of examples is set equal to that of pointwise evaluation. Results for pointwise and pairwise evaluations using gpt-4-0314 as the reference system are shown as horizontal lines. Both input sets and comparisons are sampled with replacement, with 1000 iterations.}
    \label{fig:p_pairs_sampling3}
\end{figure*}

\begin{figure*}[]
  \includegraphics[width=\linewidth]{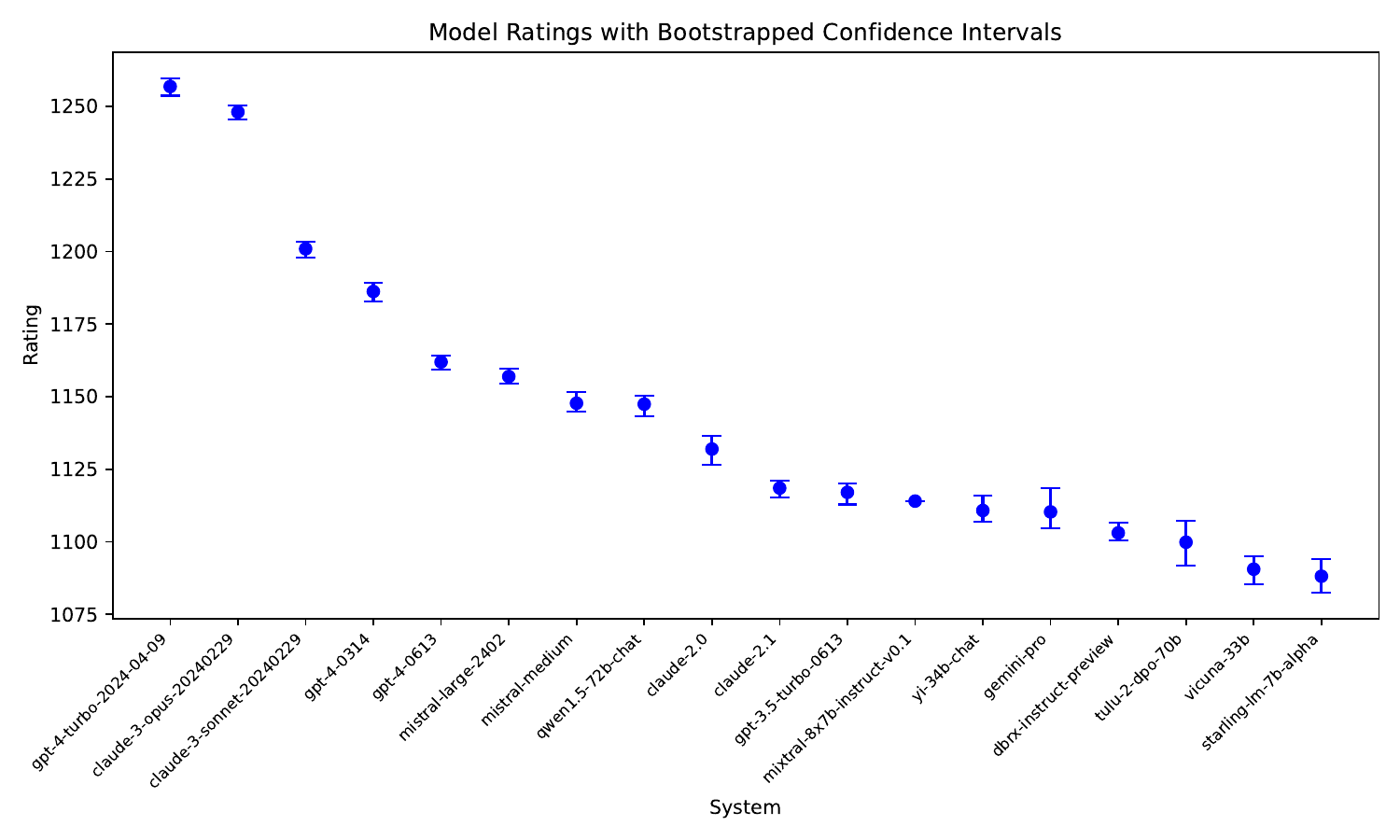}
  \caption{System ratings of Chatbot Arena with bootstrapping mean and 95\% confidence interval.}
    \label{fig:chatbot_arena_rating}
\end{figure*}

\begin{figure*}[]
  \includegraphics[width=\linewidth]{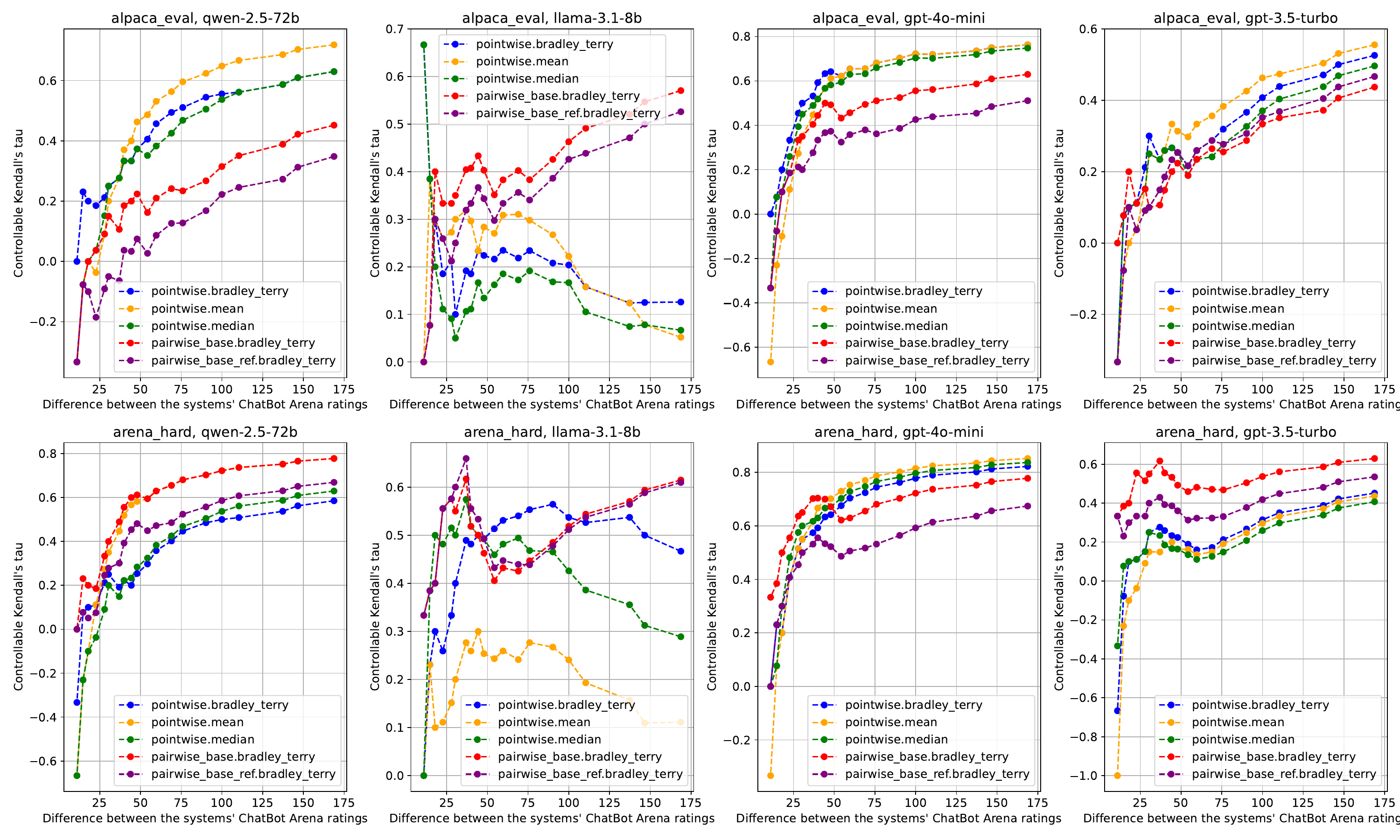}
  \caption{Controllable Kendall's tau ($\tau_u$) between the system rankings from automatic benchers and human judgment when only partial system pairs are used. The X-axis denotes the value of threshold $u$, which controls the maximum difference between the systems' ChatBot Arena ratings. Across all settings, we found that the LLM benchers' performance degrades when they evaluate close-performing systems.}
    \label{fig:controllable_kendall_tau_2}
\end{figure*}

\begin{figure*}[]
  \includegraphics[width=\linewidth]{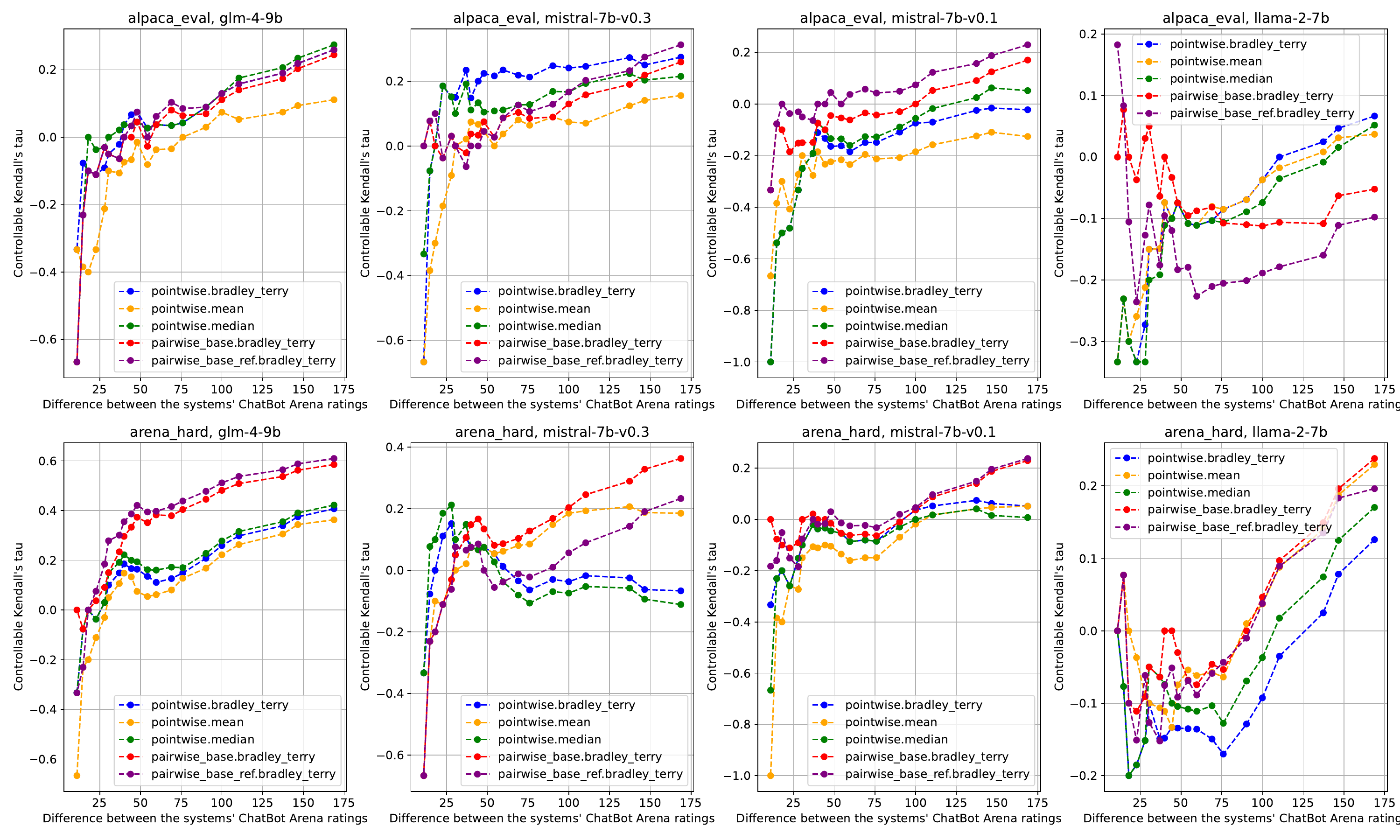}
  \caption{Controllable Kendall's tau ($\tau_u$) between the system rankings from automatic benchers and human judgment when only partial system pairs are used. The X-axis denotes the value of threshold $u$, which controls the maximum difference between the systems' ChatBot Arena ratings. Across all settings, we found that the LLM benchers' performance degrades when they evaluate close-performing systems.}
    \label{fig:controllable_kendall_tau_3}
\end{figure*}

\begin{table*}
\centering\small
\begin{tabular}{p{14cm}}
\toprule
\textbf{Prompts and Instructions} \\
\midrule
  
<|im\_start|> system\\
You are a helpful assistant in evaluating the quality of the outputs for a given instruction. Your goal is to score the output for the given instruction. \\
<|im\_end|> \\
<|im\_start|> user \\
Score the Output on a scale from 0 to 9 for the given instruction, where a score of zero means "poor quality" and score of nine means "perfect quality". The output is generated by an AI chatbot. \\
\\
Here are some rules of the evaluation: \\
(1) You should prioritize evaluating whether the output honestly/precisely/closely executes the instruction, then consider its helpfulness, accuracy, level of detail, harmlessness, etc. \\
(2) Outputs should NOT contain more/less than what the instruction asks for, as such outputs do NOT precisely execute the instruction. \\
(3) You should avoid any potential bias and your judgment should be as objective as possible. \\
\\
Do NOT provide any explanation for your choice. \\
You should answer one number from 0 to 9. Do NOT output any other words. \\
\\
\# Instruction: \\
\{INSTRUCTION\} \\
\\
\# Output: \\
\{OUTPUT\} \\
\\
\# What is your rating for the Output? \\
<|im\_end|> \\
\bottomrule
\end{tabular}
\caption{Prompt template for pointwise evaluation.}
\label{tab:prompt_pointwise}
\end{table*}

\begin{table*}
\centering\small
\begin{tabular}{p{14cm}}
\toprule
\textbf{Prompts and Instructions} \\
\midrule

<|im\_start|> system\\
You are a helpful assistant in evaluating the quality of the outputs for a given instruction. Your goal is to select the best output for the given instruction. \\
<|im\_end|> \\
<|im\_start|> user \\
Select the Output (a) or Output (b) that is better for the given instruction. The two outputs are generated by two different AI chatbots respectively. \\
\\
Here are some rules of the evaluation: \\
(1) You should prioritize evaluating whether the output honestly/precisely/closely executes the instruction, then consider its helpfulness, accuracy, level of detail, harmlessness, etc. \\
(2) Outputs should NOT contain more/less than what the instruction asks for, as such outputs do NOT precisely execute the instruction. \\
(3) You should avoid any potential bias and your judgment should be as objective as possible. For example, the order in which the outputs were presented should NOT affect your judgment, as Output (a) and Output (b) are **equally likely** to be the better. \\
\\
Do NOT provide any explanation for your choice. \\
Do NOT say both / neither are good. \\
You should answer using ONLY "Output (a)" or "Output (b)". Do NOT output any other words. \\
\\
\# Instruction: \\
\{INSTRUCTION\} \\
\\
\# Output (a): \\
\{OUTPUT\_1\} \\
\\
\# Output (b): \\
\{OUTPUT\_2\} \\
\\
\# Which is better, Output (a) or Output (b)? Your response should be either "Output (a)" or "Output (b)": \\
<|im\_end|> \\
\bottomrule
\end{tabular}
\caption{Prompt template for base pairwise evaluation.}
\label{tab:prompt_base_pairwise}
\end{table*}

\begin{table*}
\centering\small
\begin{tabular}{p{14cm}}
\toprule
\textbf{Prompts and Instructions} \\
\midrule
  
<|im\_start|> system\\
You are a helpful assistant in evaluating the quality of the outputs for a given instruction. Your goal is to select the best output for the given instruction.\\
<|im\_end|> \\
<|im\_start|> user \\
Output (a) and Output (b) are generated by two different AI chatbots for the given instruction respectively. Output one of the following choices as your verdict: \\
1. Output (a) is significantly better. \\
2. Output (a) is slightly better. \\
3. Tie, relatively the same. \\
4. Output (b) is slightly better. \\
5. Output (b) is significantly better. \\
\\
Here are some rules of the evaluation: \\
(1) You should prioritize evaluating whether the output honestly/precisely/closely executes the instruction, then consider its helpfulness, accuracy, level of detail, harmlessness, etc. \\
(2) Outputs should NOT contain more/less than what the instruction asks for, as such outputs do NOT precisely execute the instruction. \\
(3) You should avoid any potential bias and your judgment should be as objective as possible. For example, the order in which the outputs were presented should NOT affect your judgment, as Output (a) and Output (b) are **equally likely** to be the better. \\
\\
Do NOT provide any explanation for your choice. \\
Do NOT say both / neither are good. \\
You should answer using ONLY 1, 2, 3, 4, or 5. Do NOT output any other words. \\
\\
\# Instruction: \\
\{INSTRUCTION\} \\
\\
\# Output (a): \\
\{OUTPUT\_1\} \\
\\
\# Output (b): \\
\{OUTPUT\_2\} \\
\\
\# What is your verdict? Your response should be 1, 2, 3, 4, or 5: \\
<|im\_end|> \\
\bottomrule
\end{tabular}
\caption{Prompt template for 5-point pairwise evaluation.}
\label{tab:prompt_5point_pairwise}
\end{table*}

\end{document}